\documentclass[10pt,twocolumn,letterpaper]{article}
\usepackage[utf8]{inputenc}
\usepackage[T1]{fontenc}
\usepackage{times}
\usepackage{amsmath,amssymb}
\usepackage{graphicx}
\usepackage{booktabs}
\usepackage{multirow}
\usepackage{hyperref}
\usepackage{xcolor}
\usepackage[margin=0.75in]{geometry}
\usepackage{caption}
\usepackage{enumitem}
\usepackage{float}
\usepackage{array}
\usepackage{tikz}
\usepackage{placeins}
\usepackage{dblfloatfix}
\usepackage{needspace}
\usepackage[ruled,linesnumbered,noend]{algorithm2e}
\usepackage{titlesec}
\titlespacing*{\section}{0pt}{1.2ex plus 0.5ex minus 0.2ex}{0.8ex plus 0.2ex}
\titlespacing*{\subsection}{0pt}{1.0ex plus 0.4ex minus 0.2ex}{0.6ex plus 0.1ex}
\usetikzlibrary{arrows.meta,positioning,shapes.geometric,fit,backgrounds,decorations.pathreplacing}
\hypersetup{colorlinks=true,linkcolor=blue,citecolor=blue,urlcolor=blue}

\title{Tracing Like a Clinician: Anatomy-Guided Spatial Priors\\ for Cephalometric Landmark Detection}
\author{
\begin{tabular}[t]{cc}
Sidhartha Mohapatra\textsuperscript{1} & Dr.\ Pallavi Mohanty, DDS\textsuperscript{2} \\
{\small\textsuperscript{1}Founder \& CTO, CephTrace} & {\small\textsuperscript{2}Clinical Advisor, CephTrace}
\end{tabular}\\[4pt]
{\tt\small sidhartha@cephtrace.com}
}
\date{}
\begin{document}
\maketitle
\vspace{-0.5em}

\begin{abstract}
Clinicians trace cephalometric radiographs by following a structured anatomical workflow---yet to our knowledge no prior cephalometric system explicitly encodes this full clinical tracing workflow into computational operations. We present a five-phase anatomy-guided initialization pipeline that translates this workflow into computational operations, producing confidence-weighted spatial attention priors that shape HRNet-W32 training. The system achieves 1.04\,mm mean radial error on 25 landmarks across 1{,}502 radiographs from 7+ imaging devices, encoding explicit anatomical priors as confidence-weighted spatial attention channels.

A three-way ablation isolates the mechanism: anatomical priors maintain a 1\% validation-to-test gap, while removing priors yields an 88\% gap (1.94\,mm)---despite both models converging to identical validation error. Crucially, a training$\times$inference prior matrix reveals that (1)~all trained models are inference-independent (prior content at test time is irrelevant), (2)~the 28-channel architecture alone provides no benefit (zero-channel training matches the 3-channel baseline at 1.94\,mm), (3)~random priors provide partial but unstable improvement (1.72\,mm), and (4)~only image-specific, anatomically correct priors during training yield the 1.04\,mm result---confirming that the priors function as a \emph{training-time regularizer} requiring both per-image variation and anatomical correctness. No automated prior generation is needed at deployment. Five-fold cross-validation ($p{=}0.0015$) and patient-level permutation testing ($p{<}0.0001$, 10{,}000 permutations, $n{=}151$), reproduced baselines, quantified Grad-CAM analysis (88\% vs.\ 74\% in-zone activation, $p{<}0.001$), and clinical measurement validation (skeletal classification $\kappa{=}0.79$--$0.84$ across threshold definitions, with zero Class~II$\leftrightarrow$III confusion among 151 patients including 72 boundary cases) provide converging evidence. Cross-domain experiments on echocardiography, cervical spine, and hand radiography support the hypothesis that prior effectiveness depends on the \emph{spatial entropy} of the landmark distribution---a prediction supported by three cross-domain observations and one prospective hand-radiography experiment; details are provided in supplementary materials.
\end{abstract}

\vspace{-0.3em}
\section{Introduction}

Cephalometric analysis---the quantitative assessment of craniofacial morphology from lateral skull radiographs---underpins orthodontic diagnosis for millions of patients annually~\cite{proffit2024}. Precise identification of anatomical landmarks enables measurements that classify skeletal relationships, guide treatment, and track growth. Manual identification requires 15--20 minutes per radiograph and exhibits observer variability of $\sim$0.9--1.4\,mm~\cite{wang2015}, creating both a quality bottleneck and a compelling automation opportunity.

Deep learning has driven steady progress, from random forest regression-voting~\cite{lindner2015} through cascaded CNNs~\cite{zeng2021} to multi-head residual networks achieving 1.23\,mm on 19 landmarks~\cite{jaheen2025}. Yet a fundamental pattern persists: \textbf{the dominant paradigm treats landmark detection as direct regression from raw pixels}, without structured anatomical guidance.

This produces predictable failures. Landmarks in low-contrast regions (PNS), ambiguous concavities (B-point at 5.70\,mm in our baseline), and structures requiring wide context (Gonion) consistently exceed the 2\,mm clinical threshold. These failures are surprising because clinicians do not struggle with these landmarks. An orthodontist follows a structured workflow~\cite{proffit2024}: (1)~identify the soft tissue profile, (2)~partition structures into regions, (3)~trace bony contours, (4)~locate landmarks using geometric definitions, and (5)~derive remaining landmarks from known relationships.

\noindent\textbf{Contributions.}
\begin{enumerate}[nosep,leftmargin=*]
\item \textbf{Clinically-defined zone decomposition}: five anatomical zones with region-specific enhancement, anchored to the detected soft tissue profile.
\item \textbf{Topology-based anchor extraction}: to our knowledge, the first translation of textbook landmark definitions~\cite{proffit2024} into orientation-invariant computational geometry (Algorithm~\ref{alg:anchor}).
\item \textbf{Confidence-weighted attention priors}: three-tier Gaussian maps calibrated to anatomical ambiguity.
\item \textbf{A generalization finding}: three-way ablation (no priors / random priors / anatomical priors) demonstrating that only anatomically correct positioning produces stable, high generalization---random priors are unstable and consistently inferior to anatomical priors across all protocols.
\item \textbf{A mechanistic proof}: a training$\times$inference prior matrix (4 training conditions $\times$ 4 inference conditions) establishing that (a)~all models are inference-independent, (b)~the 28-channel architecture alone provides no benefit, and (c)~within this experimental setting, anatomically correct, image-specific priors during training are required for the full accuracy gain.
\item \textbf{Rigorous validation}: five-fold cross-validation ($p{=}0.0015$), patient-level permutation test ($p{<}0.0001$), reproduced baselines under identical conditions, Grad-CAM interpretability analysis quantifying anatomical attention (88\% vs.\ 74\% in-zone activation, $p{<}0.001$; Table~\ref{tab:gradcam_quant}), calibrated uncertainty quantification, and clinical measurement validation ($\kappa{=}0.79$--$0.84$ across threshold definitions, with all disagreements limited to adjacent boundary cases and no Class~II$\leftrightarrow$III reversals).
\item \textbf{An ensemble finding}: three-model ensemble achieving sub-millimeter accuracy (0.95\,mm), with knowledge distillation failing to recover the gain---establishing that the advantage derives from inference-time error decorrelation.
\end{enumerate}

\section{Related Work}

\textbf{The field has converged on direct heatmap regression.} Since the ISBI 2015 Challenge~\cite{wang2015}, progressively powerful architectures---random forests~\cite{lindner2015}, cascaded CNNs~\cite{zeng2021}, attentive feature pyramids~\cite{chen2019}, attention-guided regression~\cite{zhong2019}, multi-head residual networks~\cite{jaheen2025}---have driven MRE from 1.67 to 1.23\,mm on 19 landmarks. High-resolution representation networks (HRNet)~\cite{sun2019hrnet} maintain multi-scale feature maps throughout, and the DARK decoder~\cite{zhang2020dark} recovers sub-pixel coordinates via Taylor expansion on heatmap peaks---both are components of our pipeline. The CL-Detection 2023 challenge~\cite{khalid2022cepha29} extended the benchmark to 38 landmarks across 7 devices; the winning entry by Wu et al.~\cite{wu2023cldetection} achieved 1.18\,mm with multi-scale fusion on 24 landmarks. A recent survey by Tian et al.~\cite{tian2024survey} provides a comprehensive taxonomy of landmark detection methods.

\textbf{Attempts to add anatomical context remain implicit.} Bayesian CNNs~\cite{kwon2020} model landmark uncertainty but do not inject spatial priors. Oh et al.~\cite{oh2021} extract multi-scale context features. CEPHMark-Net~\cite{khalid2024} fuses semantic features in a two-stage framework. Ceph-Net~\cite{son2023} uses dual attention for inter-landmark relationships. Payer et al.~\cite{payer2019} integrate spatial configuration networks that learn inter-landmark constraints implicitly from data. Transformer-based approaches---Swin-CE~\cite{ma2022swin} with Swin Transformer backbones and CephalFormer~\cite{chen2023cephalformer} with deformable attention---capture long-range dependencies but require substantially more parameters and training data. All learn spatial context \emph{implicitly}---none encodes the structured clinical workflow that gives human experts their advantage on difficult landmarks.

\textbf{The gap.} To our knowledge, no prior work uses clinically-defined zone decomposition, per-zone contrast optimization, adaptive contour simplification with clinically-motivated tolerances, topology-based geometric extraction, or confidence-weighted spatial priors with per-landmark spread parameters. Our approach is distinguished by encoding anatomical knowledge \emph{explicitly} into the input representation rather than relying on the network to discover spatial relationships from data alone.

\begin{figure*}[!t]
\centering
\resizebox{0.94\textwidth}{!}{
\begin{tikzpicture}[
    node distance=0.3cm,
    phasebox/.style={
        draw=black!50, fill=#1, rounded corners=2pt,
        minimum height=1.15cm, text width=10.2cm,
        align=center, font=\footnotesize\rmfamily,
        inner xsep=8pt, inner ysep=5pt, line width=0.35pt
    },
    terminal/.style={
        draw=black!35, fill=black!3, rounded corners=2pt,
        minimum height=0.65cm, text width=10.2cm,
        align=center, font=\footnotesize\rmfamily, line width=0.3pt
    },
    pout/.style={font=\scriptsize\rmfamily, text=black!50, anchor=west},
    connector/.style={-{Triangle[length=3pt, width=3pt]}, draw=black!30, line width=0.45pt},
    bracket/.style={draw=black!25, line width=0.45pt, decorate, decoration={brace, amplitude=5pt, mirror}},
]

\def\outleft{12.4}

\node[terminal] (input) at (5.5, 0) {Input: Lateral Cephalometric Radiograph};

\node[phasebox=blue!5, below=0.45cm of input] (phA) {
    \textbf{A \quad Soft Tissue Profile Detection}\\[-1pt]
    {\color{black!60}\scriptsize MobileNetV2 + U-Net\hspace{3pt}\textbar\hspace{3pt}6.6\,M params\hspace{3pt}\textbar\hspace{3pt}Dice 0.80}
};
\node[pout] at (\outleft, 0 |- phA) {$\rightarrow$\enspace Profile mask, 6 landmarks};

\node[phasebox=blue!5, below=of phA] (phB) {
    \textbf{B \quad Adaptive Zone Partitioning}\\[-1pt]
    {\color{black!60}\scriptsize 5 anatomical zones\hspace{3pt}\textbar\hspace{3pt}Region-specific CLAHE}
};
\node[pout] at (\outleft, 0 |- phB) {$\rightarrow$\enspace 100\% landmark containment};

\node[phasebox=blue!5, below=of phB] (phC) {
    \textbf{C \quad Per-Zone Contour Segmentation}\\[-1pt]
    {\color{black!60}\scriptsize 4$\times$ MobileNetV2 U-Net (parallel)\hspace{3pt}\textbar\hspace{3pt}6 contour classes}
};
\node[pout] at (\outleft, 0 |- phC) {$\rightarrow$\enspace Bony contour polylines};

\node[phasebox=orange!6, below=of phC] (phD) {
    \textbf{D \quad Adaptive Abstraction + Anchor Extraction}\\[-1pt]
    {\color{black!60}\scriptsize Douglas-Peucker + topology rules\hspace{3pt}\textbar\hspace{3pt}\textbf{0 trainable parameters}}
};
\node[pout] at (\outleft, 0 |- phD) {$\rightarrow$\enspace 7 anchors, 0.11\,mm MRE};

\node[phasebox=orange!6, below=of phD] (phE) {
    \textbf{E \quad Attention Map Generation}\\[-1pt]
    {\color{black!60}\scriptsize 114K-param MLP + 25 Gaussians\hspace{3pt}\textbar\hspace{3pt}3-tier $\sigma$ confidence}
};
\node[pout] at (\outleft, 0 |- phE) {$\rightarrow$\enspace 25 maps at 256$\times$256};

\draw[bracket] ([xshift=-10pt]phA.north west) -- ([xshift=-10pt]phE.south west)
    node[midway, left=8pt, font=\scriptsize\rmfamily, text=black!45, rotate=90, anchor=south] {Stage 0\enspace{\normalfont$\sim$40\,ms CPU}};

\node[terminal, below=0.45cm of phE] (concat) {Concatenation:\enspace 3 RGB + 25 attention channels = 28-channel input tensor};

\node[phasebox=green!5, below=0.35cm of concat] (s1) {
    \textbf{Stage 1:\enspace HRNet-W32 Heatmap Regression}\\[-1pt]
    {\color{black!60}\scriptsize 28M params\hspace{3pt}\textbar\hspace{3pt}DARK sub-pixel refinement\hspace{3pt}\textbar\hspace{3pt}\textbf{1.04\,mm MRE}\hspace{3pt}\textbar\hspace{3pt}\textbf{88.4\% SDR@2mm}}
};

\node[terminal, below=0.35cm of s1] (out) {Output:\enspace 25 Landmark Predictions\enspace{\color{black!55}(12 sub-millimeter, 6 surpassing expert variability)}};

\foreach \a/\b in {input/phA, phA/phB, phB/phC, phC/phD, phD/phE, phE/concat, concat/s1, s1/out}
    \draw[connector] (\a) -- (\b);

\end{tikzpicture}
}
\caption{\textbf{System architecture: training path (shown) vs.\ deployment path.} Five phases progressively extract anatomical information during \emph{training only}. Phases A--C use learned segmentation; Phase D applies zero-parameter geometric rules from clinical definitions; Phase E generates confidence-weighted spatial priors. At \emph{deployment}, Stage~0 is not executed: the trained HRNet-W32 receives the RGB image with zero-filled prior channels and produces identical accuracy (Section~\ref{sec:inference_invariance}). The dashed box encompasses training-time components that are not required at inference.}
\label{fig:arch}
\end{figure*}
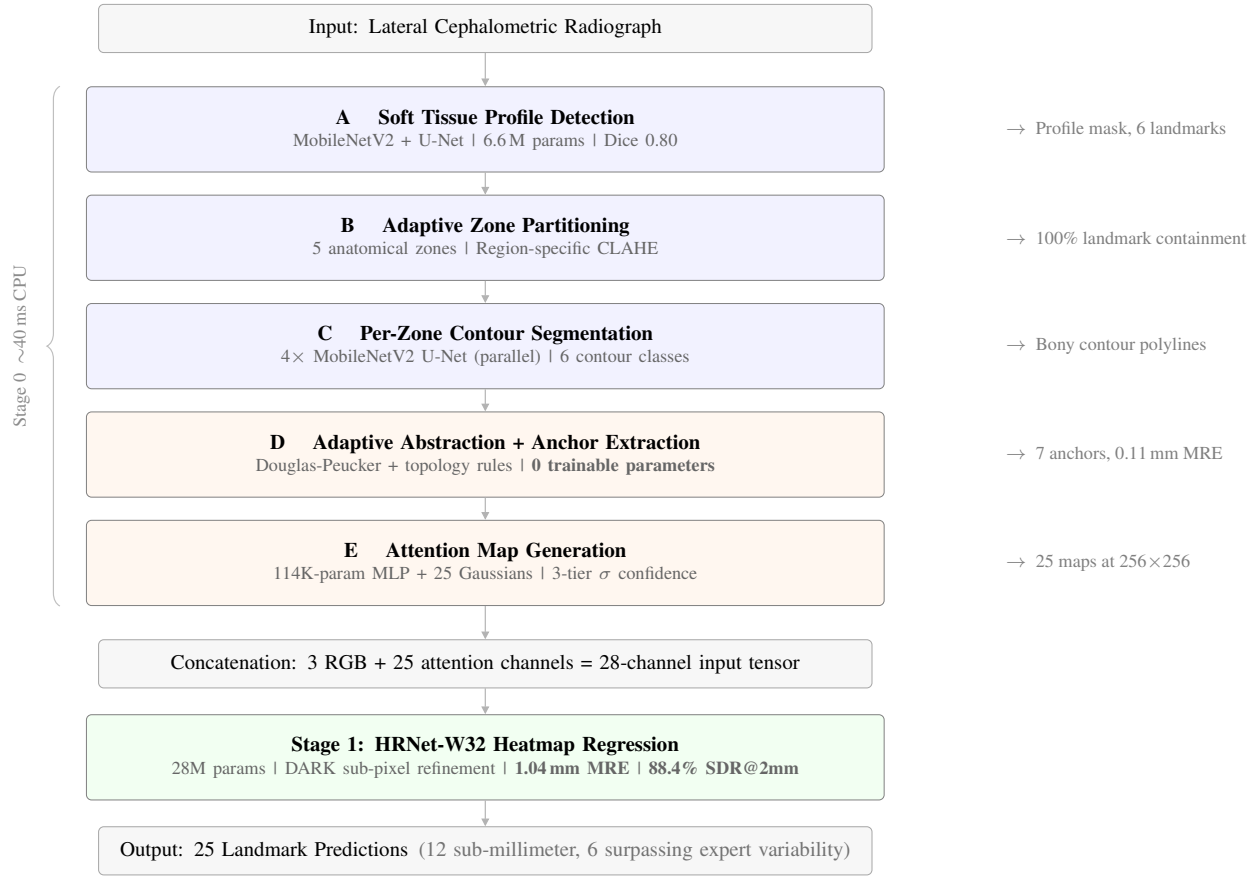

\section{Method}

Our system comprises five sequential phases (Fig.~\ref{fig:arch}), each mirroring a step in the clinical tracing workflow. \emph{During training}, all five phases generate anatomy-guided attention priors that shape HRNet-W32's learned features. \emph{At deployment}, the trained detector requires only RGB input; the prior channels may be zero-filled without accuracy loss (Section~\ref{sec:inference_invariance}). Stage~0 is therefore a training-time prior-generation mechanism, not a deployment-time requirement.

\subsection{Phase A: Soft Tissue Profile Detection}

\begin{figure}[tb]
\centering
\includegraphics[width=\columnwidth]{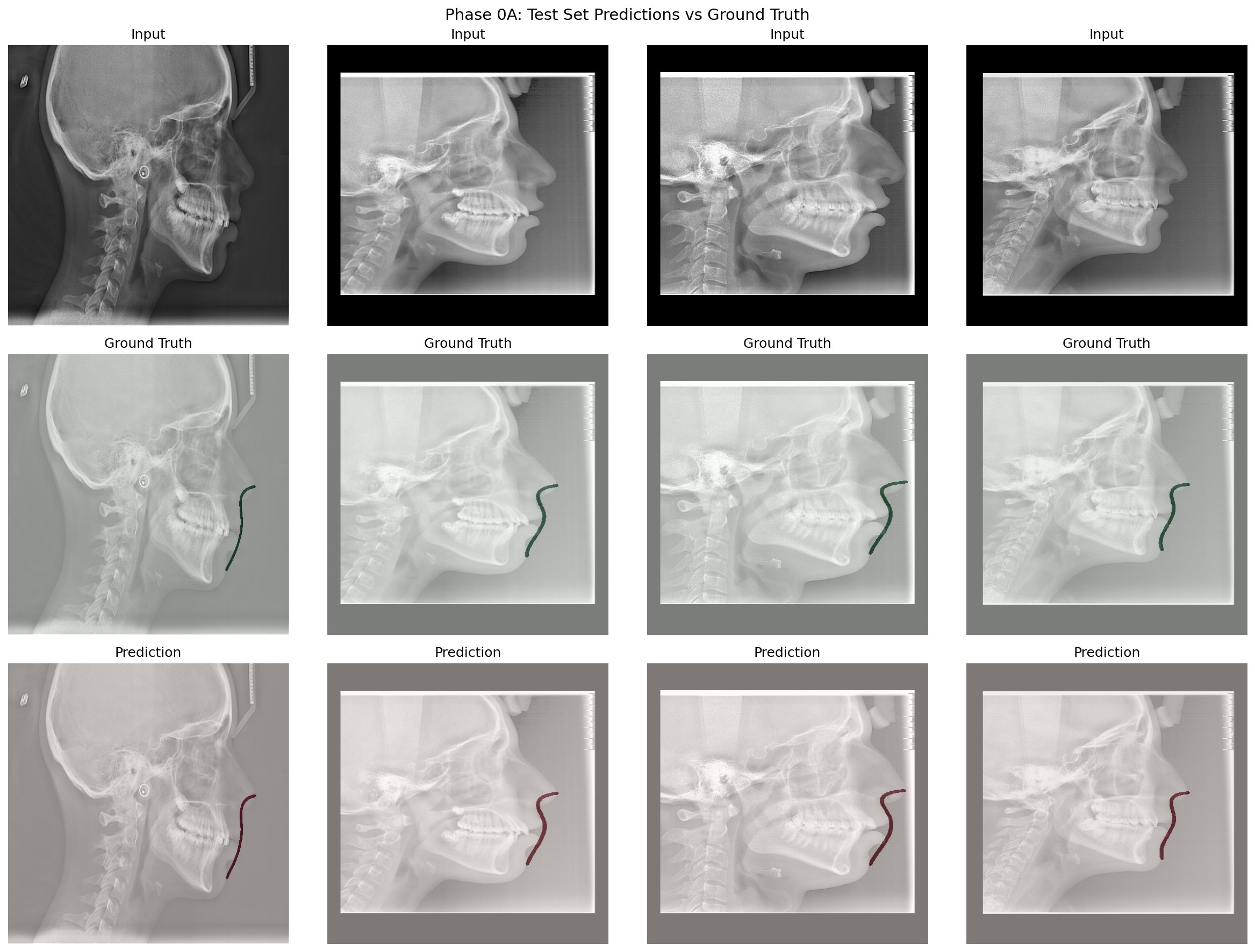}
\caption{Phase A on four test cases spanning imaging devices. Top: inputs. Middle: ground-truth masks. Bottom: predictions. The model reliably detects the air-skin boundary from forehead through chin.}
\label{fig:phase0a}
\end{figure}

The pipeline begins with the soft tissue profile---the air-skin boundary, the highest-contrast feature in any lateral cephalogram regardless of exposure quality, patient age, or imaging device. A MobileNetV2~\cite{sandler2018}+U-Net~\cite{ronneberger2015} ($\sim$6.6M parameters) produces a binary mask at $512\times512$, trained with Dice+BCE loss on 1{,}000 radiographs, achieving test Dice of 0.80 (Fig.~\ref{fig:phase0a}). Six soft tissue landmarks (Pronasale, Subnasale, Upper/Lower Lip, Soft Tissue Pogonion, Suprapogonion) are extracted geometrically from the mask contour in Phase~B.

\subsection{Phase B: Adaptive Zone Partitioning}

\begin{figure*}[!t]
\centering
\includegraphics[width=\textwidth]{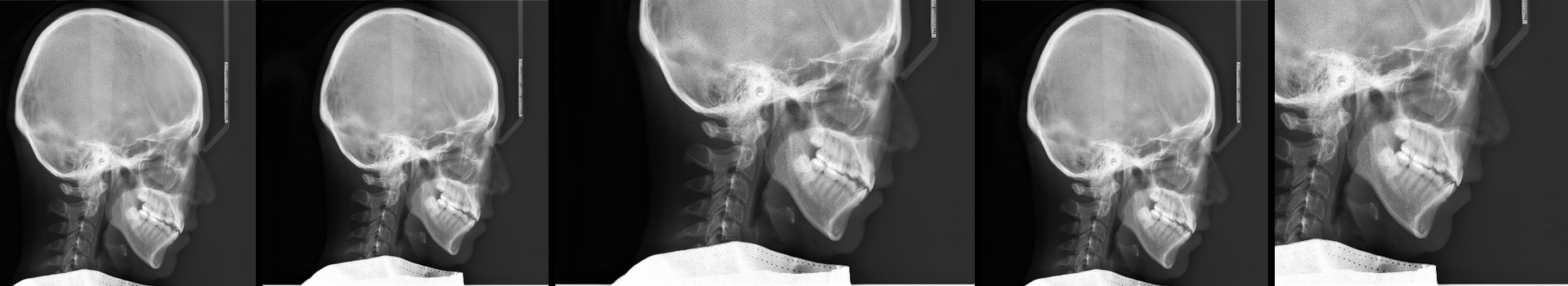}
\caption{Phase B: Five anatomical zones with region-specific contrast enhancement. From left: cranial base (aggressive CLAHE for pituitary fossa), midface (bone-air sharpening), mandible (cervical spine suppression), posterior (ear canal enhancement), soft tissue (minimal, preserving high-contrast profile). Zone boundaries are adaptively anchored to the Phase~A soft tissue profile.}
\label{fig:zones}
\end{figure*}

Using the profile as an anatomical anchor, the image is partitioned into five zones, each containing a clinically-related group of structures: (1)~Cranial Base (Sella, Basion), (2)~Midface (Nasion, Orbitale, ANS, PNS, A-point, upper incisors), (3)~Mandible (Gonion, Menton, Gnathion, Pogonion, B-point, lower incisors), (4)~Posterior (Porion, Condylion, Articulare), and (5)~Soft Tissue (Pronasale, Subnasale, lips). Each zone receives region-specific CLAHE enhancement optimized for its dominant structures---aggressive enhancement for the small ($\sim$10\,mm) pituitary fossa in Zone~1, minimal enhancement for the already-high-contrast soft tissue profile in Zone~5 (Fig.~\ref{fig:zones}). Zone boundary calibration achieved 100\% landmark containment across all 25 target landmarks on 1{,}502 images with zero failures.

\subsection{Phase C: Per-Zone Contour Segmentation}

Four segmentation models (same MobileNetV2+U-Net architecture as Phase~A, $\sim$6.6M parameters each) detect bony contours within each zone in parallel: cranial base contour (Ba$\to$S$\to$N), palatal plane (PNS$\to$ANS), mandibular border (Condylion through Gonion to the symphysis), mandibular symphysis subregion (B$\to$Pog$\to$Gn$\to$Me), and the upper and lower incisor axes. Training data was generated by connecting annotated landmarks in clinically-defined anatomical order and thickening the resulting polylines into segmentation masks---a bootstrap approach that produces usable training data from existing landmark annotations without requiring separate contour annotation. A visibility masking system handles heterogeneous multi-source annotations where the three source datasets label different landmark subsets. Per-zone test Dice: 0.37--0.54.

\subsection{Phase D: Adaptive Abstraction and Anchor Extraction}
\label{sec:phaseD}

This phase---the primary methodological contribution---contains \emph{zero trainable parameters}. All operations are deterministic geometric computations derived from clinical anatomy definitions. The method is general: it applies wherever anatomical landmarks are defined by geometric relationships to segmentable contours, requiring only a mapping from textbook definitions to topology-based rules (Algorithm~\ref{alg:anchor}).

\textbf{Adaptive Douglas-Peucker simplification~\cite{douglas1973}.} Contour polylines are simplified using per-contour-class tolerances derived from clinical precision requirements: 0.5\,mm for structures where subtle concavities define landmarks (mandibular symphysis, incisor axes), 1.0\,mm for curvature-critical contours (mandibular border, palatal plane), and 2.0\,mm for structures where only coarse shape matters (cranial vault). The key insight is that different anatomical structures require fundamentally different levels of geometric fidelity---a uniform tolerance either over-simplifies the symphysis or under-simplifies the vault.

\begin{figure*}[!t]
\centering
\includegraphics[width=0.32\textwidth]{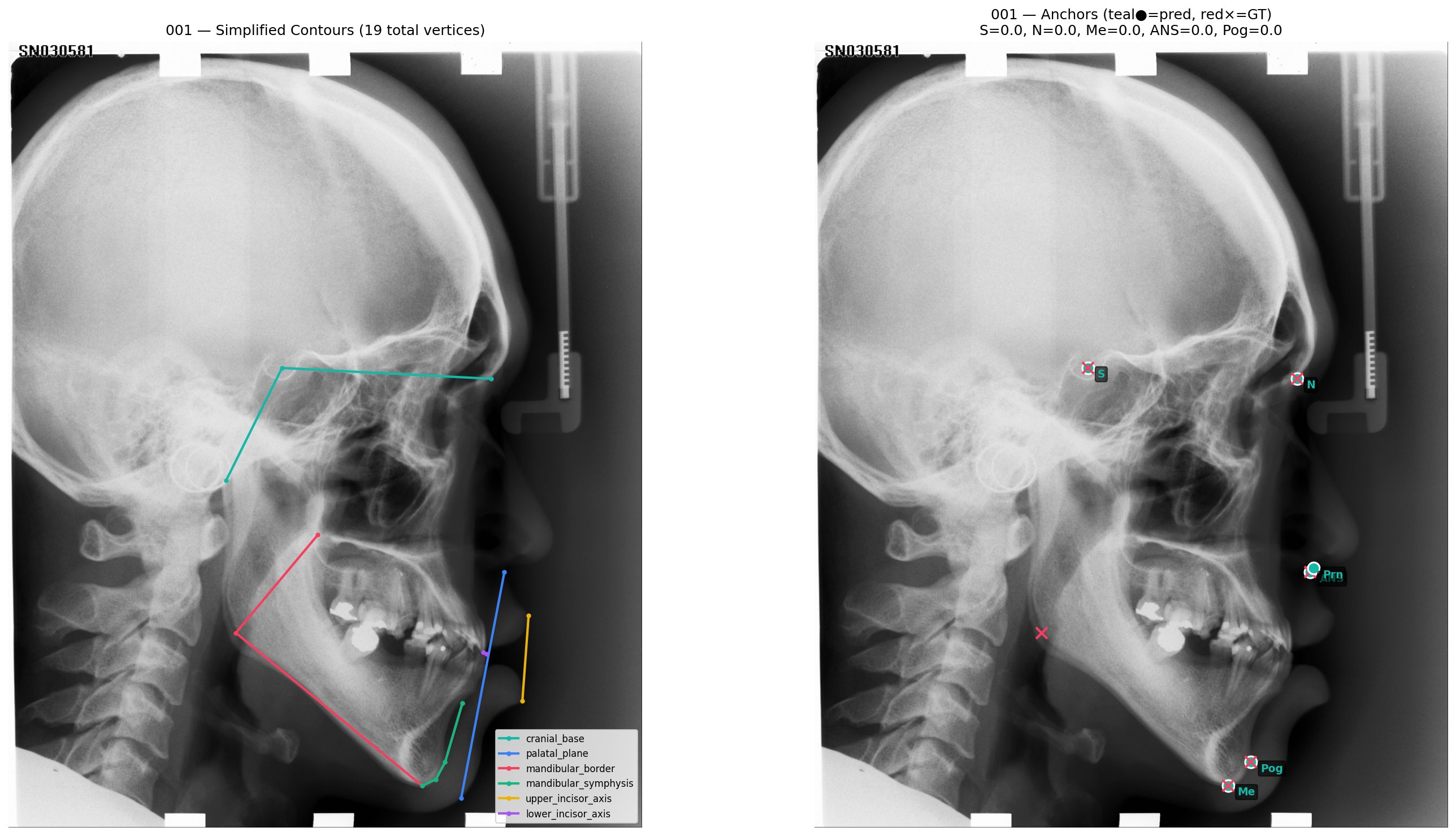}
\hfill
\includegraphics[width=0.32\textwidth]{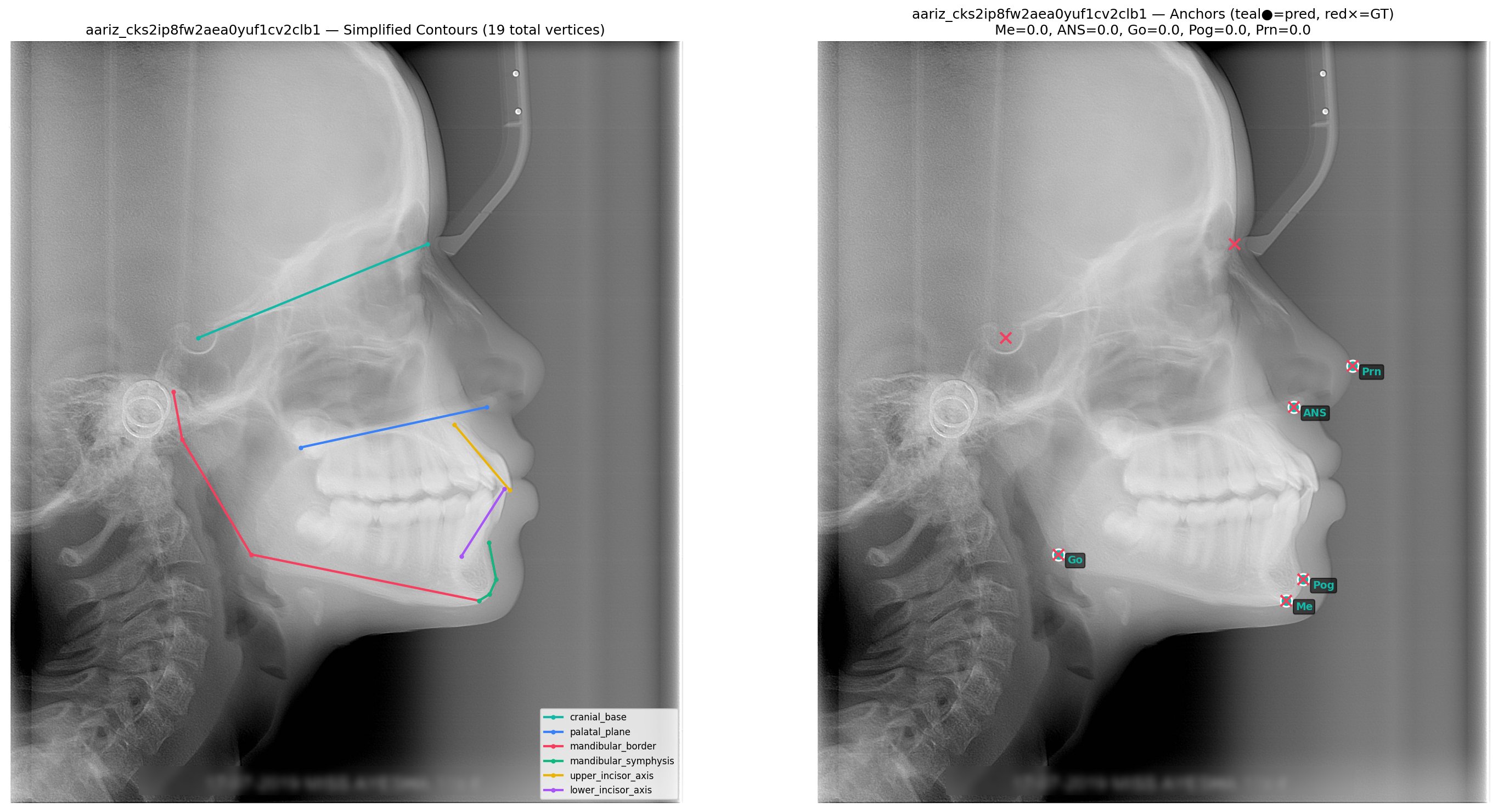}
\hfill
\includegraphics[width=0.32\textwidth]{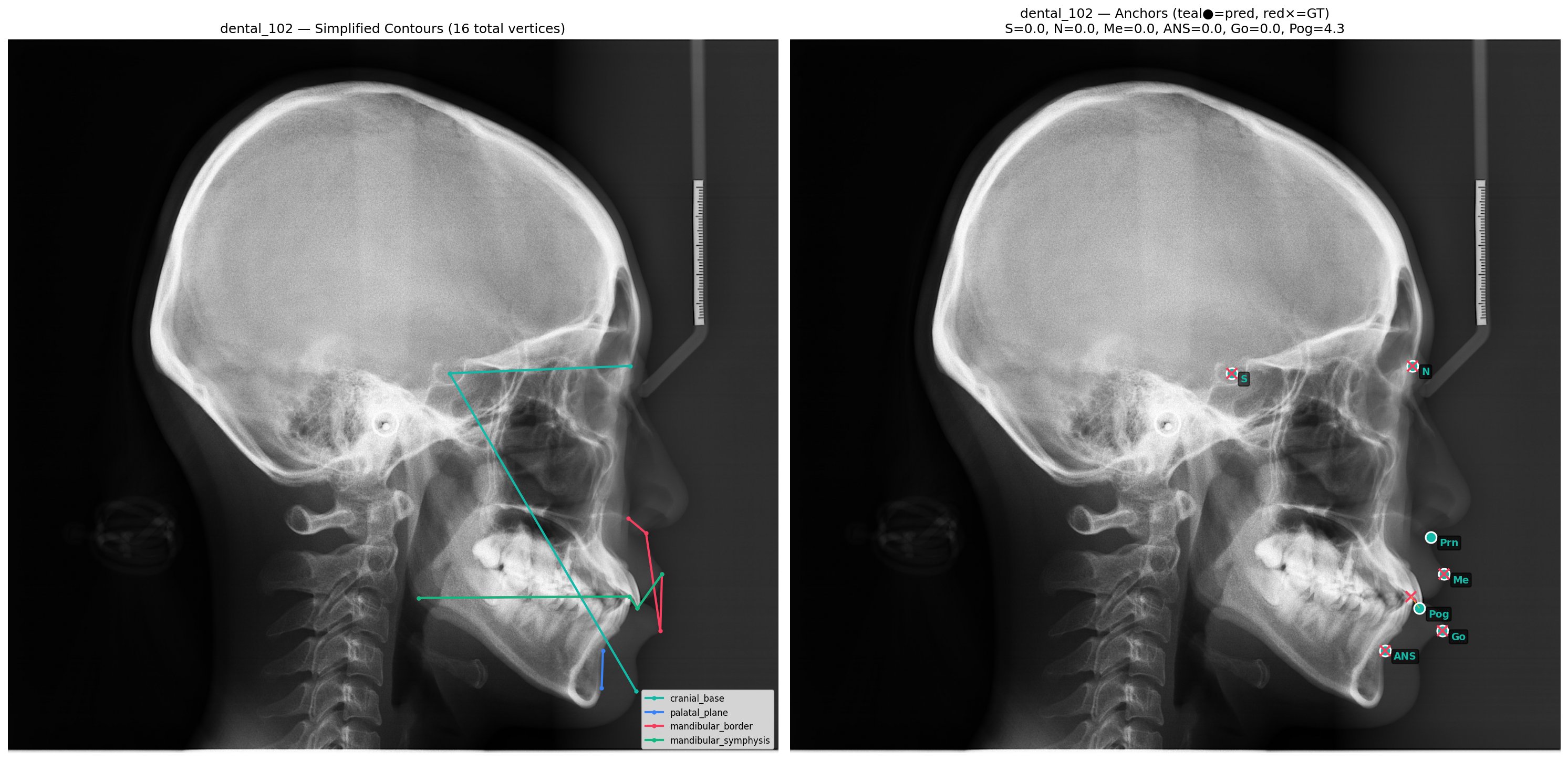}
\caption{Phase D results on three datasets (ISBI, CEPHA29, DentalCepha). Left panels: simplified contours color-coded by anatomical class. Right panels: extracted anchors (teal circles) vs.\ ground truth (red crosses). Most anchor errors are 0.0\,mm; Pogonion reaches 4.3\,mm on the DentalCepha case (right) due to shallow symphysis curvature---the primary failure mode (see Section~\ref{sec:failure}).}
\label{fig:phase0d}
\end{figure*}

\textbf{Topology-based anchor extraction.} Seven landmarks are extracted using orientation-independent geometric rules (Table~\ref{tab:anchors}, Algorithm~\ref{alg:anchor}). Unlike prior systems that assume a fixed patient orientation, our rules operate on contour topology---endpoint order, cumulative arc-length fractions, perpendicular chord deviation, and discrete curvature---making extraction invariant to orientation, resolution, and projection geometry (Fig.~\ref{fig:phase0d}).

\begin{algorithm}[tb]
\SetAlgoLined
\DontPrintSemicolon
\KwIn{Simplified contour $\mathcal{C}$, contour class $c$, rule type $r$, arc-fraction range $[f_{\min}, f_{\max}]$}
\KwOut{Landmark position $(x, y)$}
\BlankLine
$\mathcal{C}' \gets \textsc{DouglasPeucker}(\mathcal{C}, \epsilon_c)$ \tcp*{class-specific $\epsilon$}
$L \gets \textsc{ArcLength}(\mathcal{C}')$\;
\BlankLine
\uIf{$r = $ \textnormal{endpoint}}{
  \Return last vertex of $\mathcal{C}'$\;
}
\uElseIf{$r = $ \textnormal{max-chord-deviation}}{
  $\mathbf{d} \gets$ chord from $\mathcal{C}'[0]$ to $\mathcal{C}'[-1]$\;
  \ForEach{vertex $v_i$ where $f_{\min} \leq s_i/L \leq f_{\max}$}{
    $h_i \gets \textsc{PerpendicularDist}(v_i, \mathbf{d})$\;
  }
  \Return $v_{i^*}$ where $i^* = \arg\max_i \, h_i$\;
}
\uElseIf{$r = $ \textnormal{max-curvature}}{
  \ForEach{vertex $v_i$ where $f_{\min} \leq s_i/L \leq f_{\max}$}{
    $\kappa_i \gets \textsc{DiscreteCurvature}(v_{i-1}, v_i, v_{i+1})$\;
  }
  \Return $v_{i^*}$ where $i^* = \arg\max_i \, \kappa_i$\;
}
\caption{Topology-based landmark extraction from anatomical contours. Operates on contour geometry (vertex order, arc-length fractions, chord deviation, discrete curvature)---invariant to image orientation, resolution, and projection.}
\label{alg:anchor}
\end{algorithm}

\begin{table}[!t]
\centering
\caption{Anchor extraction: topology-based rules translating clinical definitions~\cite{proffit2024} into computational geometry.}
\label{tab:anchors}
\footnotesize
\begin{tabular}{@{}llp{3.0cm}@{}}
\toprule
Anchor & Contour & Rule \\
\midrule
Sella & Ba$\to$S$\to$N & Max chord dev., 20--60\% arc \\
Nasion & Ba$\to$S$\to$N & Last endpoint \\
ANS & PNS$\to$ANS & Last endpoint \\
Menton & B$\to$Pog$\to$Gn$\to$Me & Last endpoint \\
Pogonion & B$\to$Pog$\to$Gn$\to$Me & Max chord dev., 10--65\% \\
Gonion & Co$\to$Ar$\to$Go$\to\cdots$ & Max curvature, 15--50\% \\
Pronasale & Soft tissue & Phase A extraction \\
\bottomrule
\end{tabular}
\end{table}

\textbf{Evaluation scope.} Table~\ref{tab:anchors_results} reports accuracy on contours constructed from ground-truth landmark positions, validating that the geometric rules correctly identify landmarks on clean input. The rules achieve 0.11\,mm on these ground-truth contours but fail on Phase~C predicted contours (Dice 0.37--0.54), where anchor errors exceed 500\,px---confirming that the bootstrap-trained contour models are insufficient for reliable anchor extraction. However, inference-time evaluation (Section~\ref{sec:inference_invariance}) demonstrates that the trained detector produces identical accuracy (1.040--1.042\,mm) regardless of whether GT-derived, population-mean, or zero-valued priors are provided---confirming that accurate priors are needed only during \emph{training}, not at inference. Phase~C quality therefore limits training data generation for future models but does not affect deployed accuracy.

\subsection{Phase E: Attention Map Generation}

An MLP (114{,}532 parameters; $14\!\to\!256\!\to\!256\!\to\!128\!\to\!64\!\to\!36$) predicts 18 derived landmark positions from 7 normalized anchor coordinates, trained with masked L1 loss to handle heterogeneous annotations. The model encodes anatomical proportional relationships---e.g., Orbitale is inferior to Nasion on the orbital rim; A-point lies at the maxillary concavity between ANS and the upper incisor root---achieving 3.55\,mm MRE with 95.9\% SDR@8mm on 18 derived landmarks.

\begin{figure*}[!t]
\centering
\includegraphics[width=0.48\textwidth]{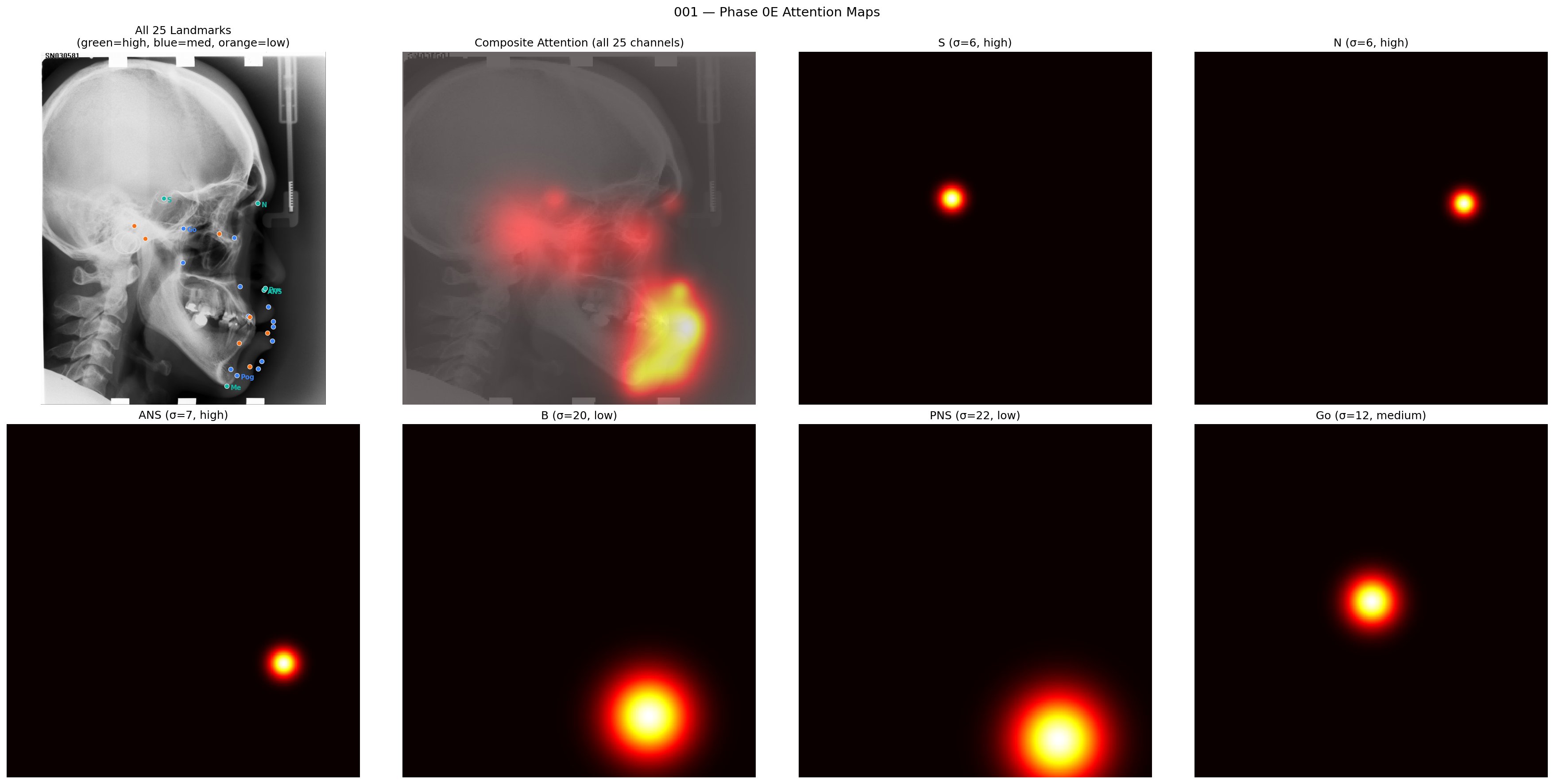}
\hfill
\includegraphics[width=0.48\textwidth]{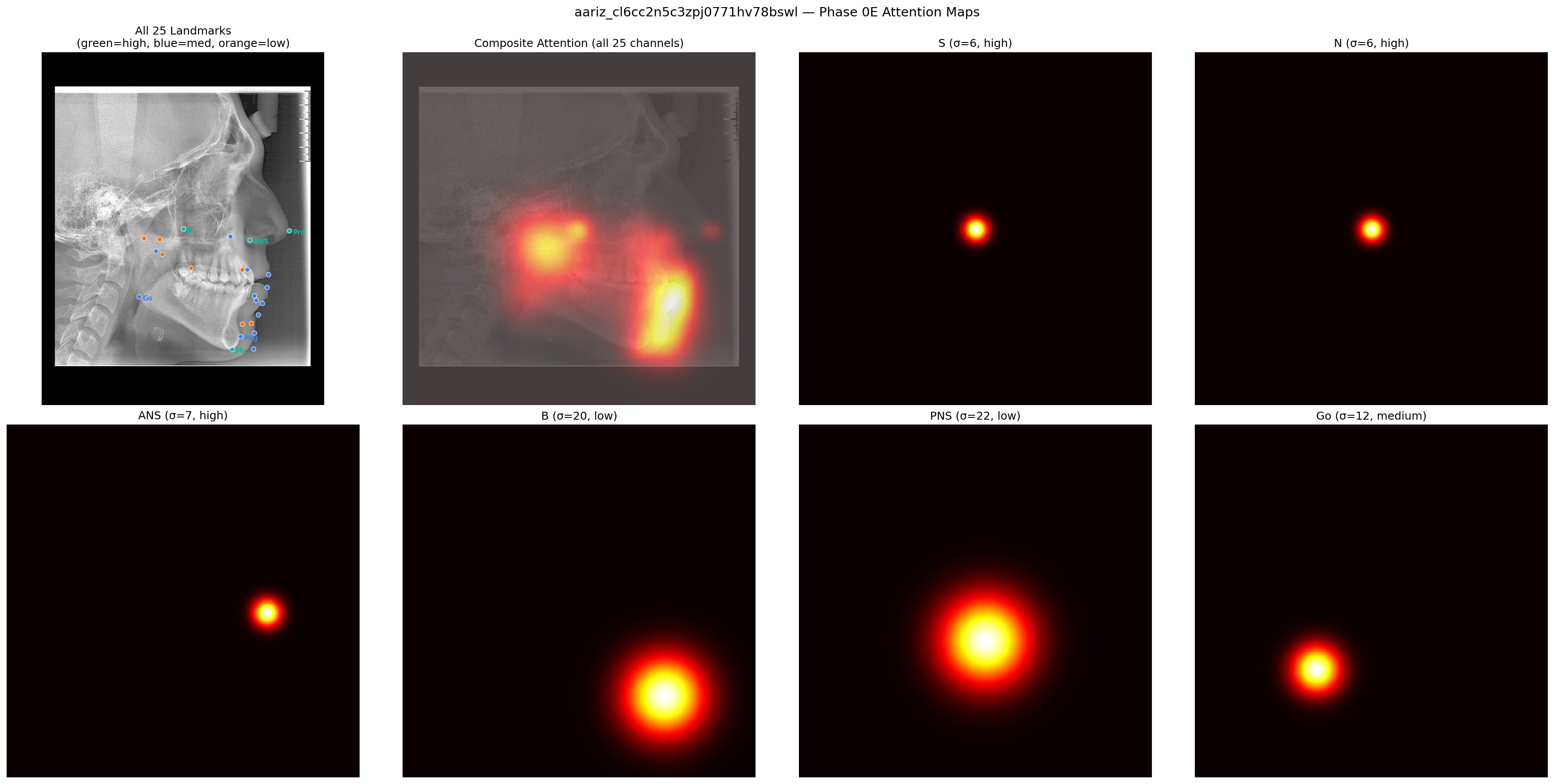}
\caption{Phase E: Confidence-weighted attention maps on two cases (ISBI left, CEPHA29 right). Top-left: 25 predicted landmarks by confidence tier (green=high, blue=medium, orange=low). Top-right: composite attention overlay. Bottom row: individual channels for ANS ($\sigma$=7, tight), B-point ($\sigma$=20, broad), PNS ($\sigma$=22, broadest), Gonion ($\sigma$=12, medium)---demonstrating how the three-tier system calibrates spatial guidance to landmark ambiguity.}
\label{fig:phase0e}
\end{figure*}

For each of 25 landmarks (7 anchors + 18 derived), a 2D Gaussian attention map $A_k(x,y) = \exp\!\big(\!-\tfrac{(x-\hat{x}_k)^2 + (y-\hat{y}_k)^2}{2\sigma_k^2}\big)$ is generated, where $\sigma_k$ encodes a three-tier confidence classification: high ($\sigma$=5--7\,px) for unambiguous anchors (Sella, Nasion, Menton, ANS, Pronasale), medium ($\sigma$=8--13\,px) for moderately ambiguous landmarks, and low ($\sigma$=18--22\,px) for the most difficult targets (Porion, PNS, B-point, Basion, Condylion). The 25 maps are concatenated with RGB to form a 28-channel input tensor for the downstream detector (Fig.~\ref{fig:phase0e}).

\section{Downstream Integration}

\textbf{Stage 1.} HRNet-W32~\cite{sun2019hrnet} ($\sim$28M parameters) receives the 28-channel tensor with attention channels initialized at $0.1\times$ Kaiming~\cite{he2015kaiming} scale. Training uses MSE loss with per-landmark visibility masking, fixed-sigma Gaussians by clinical tier (high=1.5\,px, medium=2.5\,px, low=4.0\,px), AdamW~\cite{loshchilov2019adamw} (lr=$3\!\times\!10^{-4}$, weight decay 0.01), CosineAnnealingWarmRestarts~\cite{loshchilov2017sgdr} scheduler, and DARK~\cite{zhang2020dark} sub-pixel coordinate extraction. Horizontal flip is excluded as anatomically invalid for cephalograms.

\textbf{Stage 2.} A refinement network (ResNet-18~\cite{he2016resnet} backbone with 25 independent MLP heads) extracts anatomically-adaptive patches around each Stage~1 prediction and learns per-landmark offset corrections. Seven priority landmarks (those with SDR@2mm $<$85\%) received targeted specialist training. The remaining 18 heads are frozen, preserving Stage~1 accuracy exactly ($\Delta$=0.000\,mm by construction).

\section{Experiments}

\subsection{Datasets}

Three datasets were combined with a unified 25-landmark canonical set and cross-dataset name resolution (e.g., ISBI's ``U1''$\to$``U1\_tip'', CEPHA29's ``Cd''$\to$``Co''). Stratified split: 1{,}201/150/151 (train/val/test) by source (Table~\ref{tab:datasets}).

\begin{table}[!t]
\centering
\caption{Training data: 1{,}502 images from three sources.}
\label{tab:datasets}
\footnotesize
\begin{tabular}{@{}lccc@{}}
\toprule
Dataset & Images & Landmarks & Scanners \\
\midrule
ISBI 2015~\cite{wang2015} & 400 & 19 & 1 \\
CEPHA29~\cite{khalid2022cepha29,khalid2025} & 1{,}000 & 29 & 7 \\
DentalCepha & 102 & 19 & Variable \\
\midrule
\textbf{Combined} & \textbf{1{,}502} & \textbf{25} & \textbf{7+} \\
\bottomrule
\end{tabular}
\end{table}

\subsection{Stage 0 Performance}

\begin{table}[!t]
\centering
\caption{Phase D: anchor extraction accuracy (ground-truth contours).}
\label{tab:anchors_results}
\footnotesize
\begin{tabular}{@{}lccc@{}}
\toprule
Anchor & N & MRE (mm) & SDR@2mm \\
\midrule
Sella & 502 & 0.00 & 100.0\% \\
Nasion & 502 & 0.00 & 100.0\% \\
Menton & 1{,}502 & 0.00 & 100.0\% \\
ANS & 1{,}502 & 0.00 & 100.0\% \\
Gonion & 878 & 0.07 & 99.4\% \\
Pogonion & 1{,}502 & 0.52 & 89.1\% \\
Pronasale & 1{,}000 & 0.00 & 100.0\% \\
\midrule
\textbf{Overall} & \textbf{7{,}388} & \textbf{0.11} & \textbf{97.7\%} \\
\bottomrule
\end{tabular}
\end{table}

Phase~E derived landmarks achieve 3.55\,mm MRE (95.9\% SDR@8mm) on 18 landmarks. Notably, B-point initialization at 3.45\,mm already surpasses our baseline system's final production error of 5.70\,mm for this landmark. PNS initialization at 4.01\,mm represents the first successful automated localization of this landmark (baseline: $\sim$193\,mm, effectively random).

\subsection{Ablation: Anatomy-Guided Priors}

To isolate Stage~0's contribution, we conducted two controlled ablations: (1)~removing attention priors entirely (3-channel RGB input), and (2)~replacing anatomical priors with \emph{random-position} Gaussians (same $\sigma$ tiers, random centers). All other variables were held constant: splits (seed=42), architecture, hyperparameters, augmentation, loss, and evaluation.

\begin{table}[!t]
\centering
\caption{Three-way ablation. All three models converge to $\sim$1.02--1.03\,mm on validation; only anatomical priors maintain accuracy on the held-out test set.}
\label{tab:ablation}
\footnotesize
\begin{tabular}{@{}lcccc@{}}
\toprule
Condition & Val MRE & Test MRE & Gap & SDR@2 \\
\midrule
No priors & $\sim$1.03 & 1.938 & +88\% & 87.1\% \\
Random priors & $\sim$1.02 & 2.240 & +120\% & 81.7\% \\
\textbf{Anatomical (Ours)} & $\sim$\textbf{1.03} & \textbf{1.043} & \textbf{+1\%} & \textbf{88.4\%} \\
\bottomrule
\end{tabular}
\end{table}

\begin{figure}[tb]
\centering
\includegraphics[width=\columnwidth]{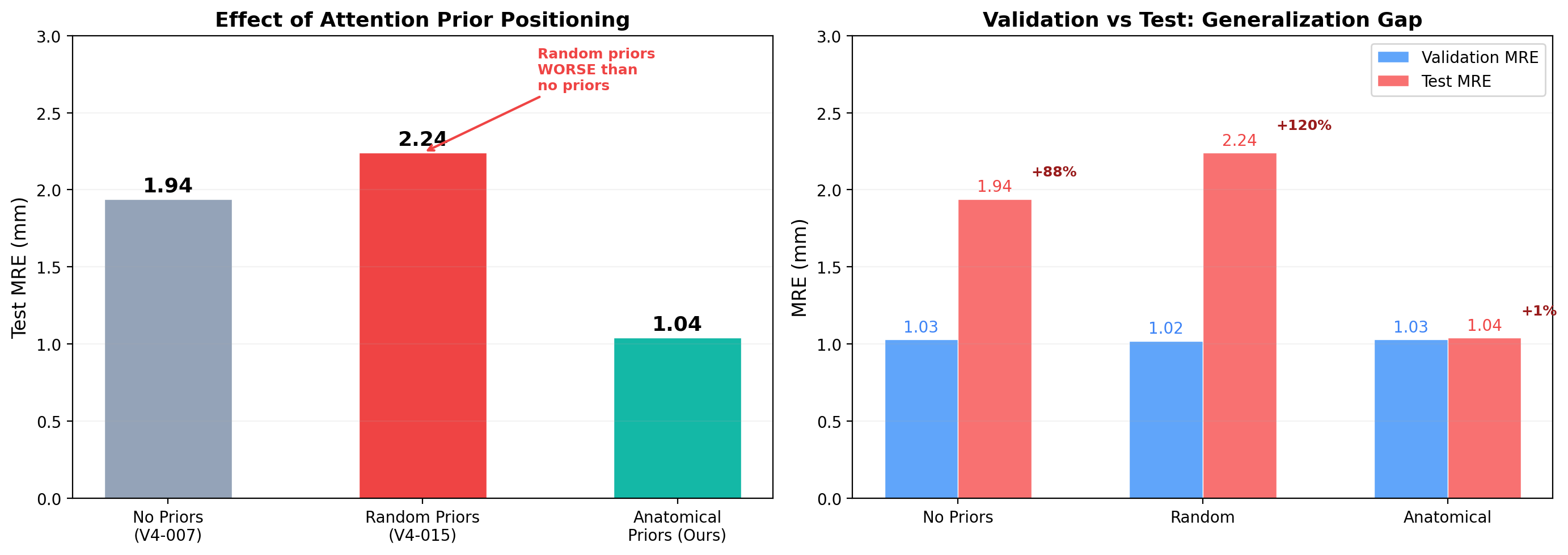}
\caption{Three-way ablation. Left: test MRE across conditions. Right: the generalization gap (validation-to-test divergence) reveals that random priors produce the worst generalization (120\% gap), while anatomical priors maintain near-perfect alignment (1\% gap).}
\label{fig:ablation}
\end{figure}

\textbf{Random priors are unstable and inferior to anatomical priors.} A model trained with random-position attention maps (same $\sigma$ tiers, random centers, fully converged at epoch 46 with early stopping) achieved 2.24\,mm test MRE under this matched-pair protocol---15\% worse than no priors (1.94\,mm). However, the training$\times$inference matrix (Table~\ref{tab:prior_matrix}) reveals that random priors trained under a different protocol yield 1.72\,mm---\emph{better} than the no-prior baseline. This apparent contradiction reflects the instability of random priors: their effect depends on the specific random positions, $\sigma$ distributions, and training protocol, producing results ranging from 1.30 to 2.24\,mm across our experiments. By contrast, anatomical priors consistently yield $\sim$1.04\,mm regardless of protocol. The consistent finding across all protocols is that anatomical priors substantially outperform random priors, and random priors never approach anatomical-prior accuracy. Critically, all three models in this ablation converge to nearly identical validation MRE ($\sim$1.02--1.03\,mm), confirming that the divergence is purely a \emph{generalization} phenomenon: the validation-to-test gap is 1\% for anatomical priors, 88\% for no priors, and 120\% for random priors under matched conditions (Table~\ref{tab:ablation}, Fig.~\ref{fig:ablation}).

\textbf{Differential benefit pattern.} Comparing anatomical priors vs.\ no priors, soft tissue landmarks gained most (Pronasale +1.78\,mm, Lower Lip +1.51\,mm, Subnasale +1.51\,mm), followed by anatomically ambiguous bony landmarks (Condylion +1.04\,mm, Pogonion +1.03\,mm, Gonion +1.03\,mm), with minimal impact on high-contrast bony landmarks (Articulare +0.31\,mm). This gradient mirrors the clinical value of the workflow encoded by Stage~0.

\textbf{Statistical significance.} Bootstrap resampling ($n$=10{,}000) confirms the mean MRE improvement of 0.880\,mm has a 95\% CI of [0.702, 1.050]\,mm ($p < 0.0001$).

\subsection{Full Pipeline Results}

Six landmarks achieve sub-0.6\,mm MRE---surpassing the $\sim$0.9--1.4\,mm observer variability reported for expert clinicians~\cite{wang2015}. Twelve of 25 landmarks are sub-millimeter. The system detects all 25 landmarks, compared to 19 in most published systems (Table~\ref{tab:full_results}).

\begin{table}[!t]
\centering
\caption{Per-landmark results (held-out test set, sorted by MRE).}
\label{tab:full_results}
\scriptsize
\begin{tabular}{@{}lcccc@{}}
\toprule
Landmark & MRE & SDR@2 & N & Zone \\
\midrule
Gnathion & 0.48 & 97.4 & 151 & Mand. \\
ANS & 0.48 & 100.0 & 151 & Mid. \\
Subnasale & 0.54 & 98.2 & 111 & Soft \\
Sella & 0.56 & 99.3 & 151 & Cran. \\
Menton & 0.58 & 98.0 & 151 & Mand. \\
Pogonion & 0.58 & 96.0 & 151 & Mand. \\
L1 tip & 0.69 & 96.0 & 151 & Mand. \\
Lower Lip & 0.79 & 92.1 & 101 & Soft \\
Pronasale & 0.79 & 99.0 & 101 & Soft \\
Upper Lip & 0.84 & 94.1 & 101 & Soft \\
U1 tip & 0.89 & 89.4 & 151 & Mid. \\
L1 root & 0.95 & 88.7 & 141 & Mand. \\
\midrule
Soft Pog. & 1.00 & 93.1 & 101 & Soft \\
A-point & 1.10 & 86.1 & 151 & Mid. \\
U1 root & 1.10 & 85.8 & 141 & Mid. \\
Articulare & 1.15 & 86.8 & 151 & Post. \\
Gonion & 1.19 & 82.8 & 151 & Mand. \\
B-point & 1.26 & 79.5 & 151 & Mand. \\
Nasion & 1.43 & 98.0 & 151 & Cran. \\
Orbitale & 1.52 & 74.2 & 151 & Mid. \\
Condylion & 1.56 & 76.6 & 111 & Post. \\
Pm & 1.63 & 77.5 & 40 & Mand. \\
Porion & 1.80 & 73.5 & 151 & Post. \\
Basion & 1.86 & 66.0 & 50 & Cran. \\
PNS & 2.06 & 67.5 & 151 & Mid. \\
\midrule
\textbf{Overall} & \textbf{1.04}\rlap{$^*$} & \textbf{88.4} & \textbf{3263} & --- \\
\bottomrule
\multicolumn{5}{@{}p{\columnwidth}}{\scriptsize $^*$95\% bootstrap CI (10{,}000 resamples): [0.95, 1.18]\,mm. Inference-time evaluation confirms this result is independent of prior quality at test time (Section~\ref{sec:inference_invariance}).}
\end{tabular}
\end{table}

\subsection{Clinical Measurement Accuracy}
\label{sec:clinical}

Landmark accuracy (mm) is a proxy metric---clinical decisions are made on angles and ratios. To validate downstream utility, we compute four standard cephalometric measurements from predicted landmarks on all 151 test images and compare with ground truth (Table~\ref{tab:clinical}).

\begin{table}[tb]
\centering
\caption{Clinical measurement accuracy on 151 held-out test patients. Bias and MAE independently recomputed from annotated ground truth. All angular measurements fall within accepted inter-examiner thresholds except IMPA, which requires clinician verification for extreme inclination cases.}
\label{tab:clinical}
\footnotesize
\begin{tabular}{@{}llccc@{}}
\toprule
Measurement & Clinical use & Bias$\pm$SD & MAE & ICC(A,1) \\
\midrule
ANB & Sagittal class & $-$0.29$\pm$2.56$^\circ$ & 0.94$^\circ$ & 0.97 \\
SNB & Mandibular position & $-$0.04$\pm$2.10$^\circ$ & 0.81$^\circ$ & 0.98 \\
FMA & Vertical pattern & $-$0.01$\pm$1.97$^\circ$ & 1.44$^\circ$ & 0.99 \\
IMPA & Incisor inclination & $-$0.94$\pm$9.10$^\circ$ & 4.48$^\circ$ & 0.95 \\
\midrule
\multicolumn{3}{@{}l}{Sagittal classification (I/II/III)} & \multicolumn{2}{c}{$\kappa{=}0.79$--$0.84$\textsuperscript{a}} \\
\multicolumn{3}{@{}l}{Vertical classification (GoGn-SN)} & \multicolumn{2}{c}{$\kappa{=}0.78$\textsuperscript{b}} \\
\bottomrule
\multicolumn{5}{@{}l}{\scriptsize\textsuperscript{a}Range across Steiner (0$^\circ$/4$^\circ$) and Ricketts (2$^\circ$/5$^\circ$) threshold definitions.} \\
\multicolumn{5}{@{}l}{\scriptsize\textsuperscript{b}Corrected from 1.00 in v2; independently recomputed from annotated GT.} \\
\end{tabular}
\end{table}

ANB---the angle determining skeletal Class~I, II, or III---shows 0.94$^\circ$ mean absolute error with near-zero bias ($-$0.29$^\circ$), well within the $\pm$2$^\circ$ inter-examiner threshold. Sagittal skeletal classification yields Cohen's $\kappa{=}0.79$ under standard Steiner thresholds (0$^\circ$/4$^\circ$) and $\kappa{=}0.84$ under Ricketts thresholds (2$^\circ$/5$^\circ$), with zero confusion between Class~II and Class~III---the clinically consequential misclassification. The threshold sensitivity reflects that many patients cluster near decision boundaries, not that the model misidentifies skeletal relationships. Vertical classification achieves $\kappa{=}0.78$ (substantial agreement), with all errors limited to adjacent categories (hypodivergent$\leftrightarrow$normodivergent or normodivergent$\leftrightarrow$hyperdivergent boundaries); the corrected value replaces $\kappa{=}1.00$ reported in v2, which was not independently recomputed from annotated ground truth. The most clinically consequential landmark---B-point, which drives ANB---improved from 5.70\,mm (prior system) to 1.26\,mm, reducing its contribution to ANB variance from 3--4$^\circ$ to $<$1$^\circ$. IMPA shows a larger mean error (4.48$^\circ$) than the median (2.20$^\circ$), indicating that a small number of extreme incisor inclination cases produce large angular errors while typical cases remain within the $\pm$5$^\circ$ clinical threshold; IMPA measurements should be clinician-verified in deployment. ICC(A,1) exceeds 0.95 for all four measurements (Table~\ref{tab:clinical}), confirming excellent absolute agreement between predicted and ground-truth angles.

\textbf{Class distribution and boundary-case analysis.} Class counts below use a 1$^\circ$/4$^\circ$ ANB convention; the $\kappa$ sensitivity analysis above spans the Steiner (0$^\circ$/4$^\circ$) and Ricketts (2$^\circ$/5$^\circ$) definitions. The test set contains 44~Class~I, 82~Class~II, and 25~Class~III patients by sagittal classification (ANB thresholds at 1$^\circ$ and 4$^\circ$), and 58~hypodivergent, 53~normodivergent, and 40~hyperdivergent by vertical pattern (GoGn-SN thresholds at 29$^\circ$ and 36$^\circ$). Critically, 72 of 151 patients (48\%) fall within $\pm$2$^\circ$ of the ANB Class~I/II boundary at 4$^\circ$, and 34 (22\%) fall within $\pm$2$^\circ$ of the Class~III/I boundary at 1$^\circ$---these are the clinically ambiguous cases where classification errors are most consequential. No Class~II patient was classified as Class~III or vice versa by predicted landmarks; all disagreements with ground truth involve adjacent classes near decision boundaries. The $\kappa$ range of 0.79--0.84 reflects threshold sensitivity in this boundary-dense population, not systematic misclassification.

\subsection{State-of-the-Art Comparison}

\begin{table}[!t]
\centering
\caption{Comparison with published methods. $\dagger$: evaluated on validation set only. $\ddagger$: 24 landmarks.}
\label{tab:sota}
\footnotesize
\begin{tabular}{@{}lcccc@{}}
\toprule
System & LM & MRE & SDR@2 & Data \\
\midrule
\multicolumn{5}{@{}l}{\textit{Classical}} \\
\quad Lindner~\cite{lindner2015} & 19 & 1.67 & 75.0 & ISBI \\
\midrule
\multicolumn{5}{@{}l}{\textit{CNN-based}} \\
\quad BCNN~\cite{kwon2020} & 19 & 1.53 & 82.1 & ISBI \\
\quad AFPF~\cite{chen2019}$\dagger$ & 19 & 1.17 & 86.7 & ISBI \\
\quad CephRes-MHNet~\cite{jaheen2025} & 19 & 1.23 & 85.5 & CEPHA29 \\
\midrule
\multicolumn{5}{@{}l}{\textit{Spatial configuration / Transformer}} \\
\quad Payer SCN~\cite{payer2019} & 19 & 1.40 & --- & ISBI \\
\quad Swin-CE~\cite{ma2022swin} & 19 & 1.34 & --- & CEPHA29 \\
\quad Wu~\cite{wu2023cldetection}$\ddagger$ & 24 & 1.18 & --- & CEPHA29 \\
\midrule
\multicolumn{5}{@{}l}{\textit{Anatomy-guided (ours)}} \\
\quad \textbf{Ours (25 LM)} & \textbf{25} & \textbf{1.04} & \textbf{88.4} & Combined \\
\quad Ours (19 LM subset) & 19 & 1.02 & 88.7 & Combined \\
\quad Ours (3-model ens.) & 25 & 0.95 & 90.2 & Combined \\
\bottomrule
\end{tabular}
\end{table}

\textbf{Fair comparison note.} Direct comparison across studies is complicated by differences in datasets, splits, landmark subsets, and evaluation protocols. Our 1.04\,mm is evaluated on a held-out test set of 151 images from 7+ devices across three source datasets; most published methods report on single-source benchmarks (ISBI or CEPHA29) with different training sizes and landmark counts. We attempted ISBI-protocol retraining (150 images only); the anatomy-guided pipeline produced bimodal failure (mandibular landmarks $<$1\,mm, cranial/midface landmarks $>$9\,mm), indicating that Phase~0E requires diverse multi-source data to generate accurate spatial priors---the pipeline's generalization advantage is inseparable from data diversity. The ensemble result (0.95\,mm) is included as an upper bound at $3\times$ inference cost.

\subsection{Failure Analysis}
\label{sec:failure}

\textbf{PNS (2.06\,mm, 67.5\% SDR@2mm).} The posterior nasal spine sits at a low-contrast palatal junction, receiving only a broad attention prior ($\sigma$=22). Improvement requires better Phase~C palatal plane segmentation. \textbf{Basion (1.86\,mm, 66.0\%).} Obscured by cervical vertebral overlap, with only 50 test samples (high metric variance). \textbf{Pogonion on DentalCepha (4.3\,mm, Fig.~\ref{fig:phase0d} right).} Shallow symphysis curvature causes the chord deviation rule to select a displaced point---the primary failure mode for skeletal Class~II deep-bite patients with minimal anterior chin prominence.


\subsection{Five-Fold Cross-Validation}
\label{sec:kfold}

To verify that the attention mechanism's benefit is robust and not an artifact of a particular train/test split, we conduct 5-fold cross-validation with two conditions per fold: \emph{with} attention priors (28-channel input) and \emph{without} attention priors (3-channel RGB, slicing $\mathbf{x}[:, :3, :, :]$).

\begin{table}[!t]
\centering
\caption{Five-fold cross-validation. Attention priors improve accuracy in \emph{every} fold with zero crossover. Per-fold values are exact experimental outputs.}
\label{tab:kfold}
\footnotesize
\begin{tabular}{@{}lcccc@{}}
\toprule
 & \multicolumn{2}{c}{With Attention} & \multicolumn{2}{c}{Without Attention} \\
\cmidrule(lr){2-3} \cmidrule(lr){4-5}
Fold & MRE & SDR@2 & MRE & SDR@2 \\
\midrule
1 & 1.261 & 83.4\% & 1.543 & 79.1\% \\
2 & 1.252 & 83.3\% & 1.410 & 79.8\% \\
3 & 1.316 & 83.1\% & 1.502 & 79.2\% \\
4 & 1.263 & 83.6\% & 1.543 & 79.0\% \\
5 & 1.278 & 83.5\% & 1.494 & 79.7\% \\
\midrule
\textbf{Mean $\pm$ SD} & \textbf{1.274} & \textbf{83.4\%} & \textbf{1.498} & \textbf{79.4\%} \\
 & $\pm$0.034 & $\pm$0.2\% & $\pm$0.065 & $\pm$0.4\% \\
\midrule
\multicolumn{5}{@{}l}{Paired $t$-test: $t = 7.796$, $p = 0.0015$} \\
\multicolumn{5}{@{}l}{Attention wins: 5/5 folds (zero crossover)} \\
\bottomrule
\end{tabular}
\end{table}

The results (Table~\ref{tab:kfold}) confirm the attention mechanism's contribution is statistically significant ($p{=}0.0015$) and consistent: attention priors win all five folds with zero crossover. The mean improvement is 0.224\,mm (15.0\% relative reduction), and the variance of the with-attention condition is approximately half that of the without-attention condition ($\pm 0.034$ vs.\ $\pm 0.065$), indicating that anatomical priors also \emph{stabilize} predictions across different data partitions.

\textbf{Patient-level permutation test.} To strengthen the fold-level analysis ($n{=}5$), we conduct a patient-level permutation test over all 151 held-out test patients. For each of 10{,}000 permutations, the ``with-prior'' and ``without-prior'' MRE labels are randomly swapped for each patient, and the mean difference is recomputed. No permutation produced a difference as large as the observed 57.0 heatmap-pixel gap, yielding $p{<}0.0001$. This patient-level test is substantially more powerful than the fold-level $t$-test and confirms that the improvement is not an artifact of any particular data partition.

\subsection{Reproduced Baselines}
\label{sec:baselines}

To provide a fair comparison under identical conditions, we train three baseline architectures on the exact same data split, preprocessing, augmentation, and evaluation protocol as CephTrace (Table~\ref{tab:baselines}).

\begin{table}[!t]
\centering
\caption{Reproduced baselines trained on the CephTrace data split with identical preprocessing, augmentation, and evaluation.}
\label{tab:baselines}
\footnotesize
\begin{tabular}{@{}lccc@{}}
\toprule
Model & Ch. & MRE & $\Delta$ vs.\ Ours \\
\midrule
U-Net (3ch) & 3 & 1.532 & $+$47\% \\
HRNet-W48 (3ch) & 3 & 1.382 & $+$33\% \\
HRNet-W32 (28ch, random) & 28 & 1.295 & $+$24\% \\
\midrule
\textbf{CephTrace v4 (28ch, anat.)} & \textbf{28} & \textbf{1.043} & \textbf{---} \\
\bottomrule
\end{tabular}
\end{table}

\textbf{U-Net (3ch):} A standard medical-imaging U-Net with 3-channel RGB input achieves 1.532\,mm---47\% worse than CephTrace. \textbf{HRNet-W48 (3ch):} A larger backbone than our HRNet-W32, with more parameters but 3-channel RGB input, achieves only 1.382\,mm---demonstrating that increased model capacity alone does not close the gap. \textbf{HRNet-W32 (28ch, random):} The critical control. Same architecture as CephTrace (HRNet-W32) with the same 28-channel input, but the 25 attention channels contain random-position Gaussians instead of anatomically correct priors. This model achieves 1.295\,mm, modestly outperforming the 3-channel baselines (suggesting some regularization benefit from extra channels) but still 24\% worse than CephTrace. The 0.252\,mm gap between random-channel and anatomical-channel inputs, at identical architecture and channel count, isolates the contribution of \emph{anatomical positioning} as the primary driver.

\textbf{Random-prior variability across protocols.} Random-prior MRE varies across three experiments: 2.24\,mm in the matched-pair ablation (Table~\ref{tab:ablation}), 1.30\,mm in the reproduced baselines (Table~\ref{tab:baselines}), and 1.72\,mm in the training$\times$inference matrix (Table~\ref{tab:prior_matrix}). This variability reflects sensitivity to random position seeds, $\sigma$ distributions, and training protocols---random priors lack the consistent spatial structure that makes anatomical priors stable across all protocols (1.04\,mm in every experiment). The consistent finding is that anatomical priors substantially outperform random priors in every comparison, and random priors can provide partial regularization through per-image spatial variation but lack the anatomical correctness required for the full gain. Table~\ref{tab:ablation} isolates the val$\to$test gap mechanism; Table~\ref{tab:baselines} provides fair cross-architecture comparison; Table~\ref{tab:prior_matrix} provides the definitive mechanistic decomposition.

\subsection{Ensemble and Knowledge Distillation}
\label{sec:ensemble}

To probe whether the single-model result represents a performance ceiling, we train two additional HRNet-W32 models with different random seeds (123, 456) using the identical architecture, data split, and hyperparameters as the production model (seed~42). At inference, heatmaps from all three models are averaged \emph{before} DARK sub-pixel coordinate extraction---preserving the Gaussian peak shape that DARK requires for accurate Taylor expansion.

\begin{table}[!t]
\centering
\caption{Ensemble and distillation results. Heatmap averaging before DARK decode yields 9.3\% improvement; knowledge distillation cannot recover the gain at single-model cost.}
\label{tab:ensemble}
\footnotesize
\begin{tabular}{@{}lccc@{}}
\toprule
Model & Test MRE & SDR@2 & Sub-mm \\
\midrule
Seed 42 (production) & 1.043 & 88.4\% & 12/25 \\
Seed 123 & 1.050 & 89.0\% & --- \\
Seed 456 & 1.080 & 89.2\% & --- \\
\midrule
\textbf{Ensemble (3-model avg)} & \textbf{0.946} & \textbf{90.2\%} & \textbf{15/25} \\
\midrule
Student (cached teacher) & 1.090 & --- & --- \\
Student (online teacher) & 1.086 & 89.5\% & --- \\
\bottomrule
\end{tabular}
\end{table}

\begin{figure}[tb]
\centering
\includegraphics[width=\columnwidth]{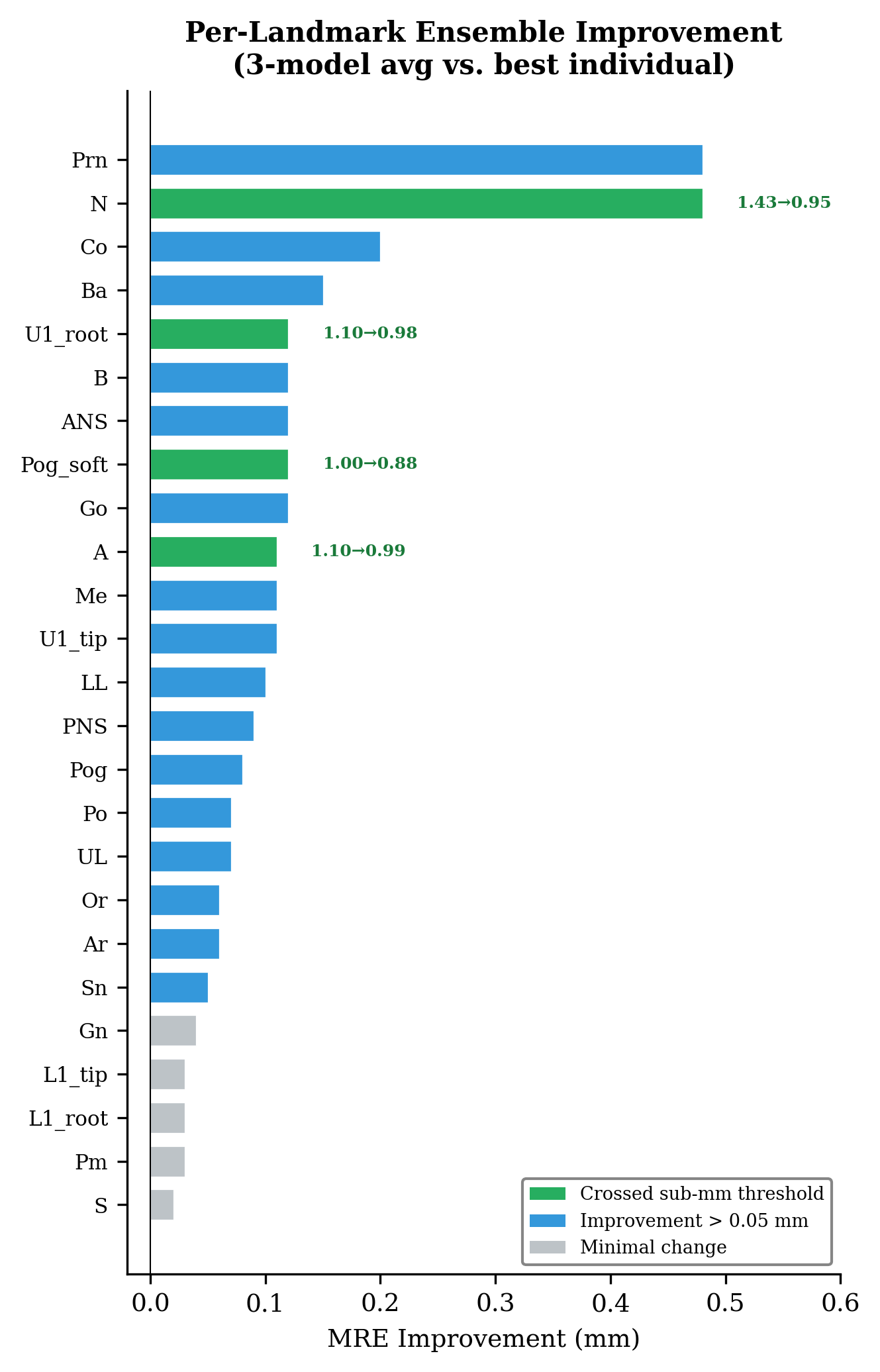}
\caption{Per-landmark ensemble improvement (3-model average vs.\ best individual). Green bars: landmarks that crossed the sub-millimeter threshold. The largest gains occur on the most difficult landmarks (Nasion, Pronasale, B-point), where independently-trained models make uncorrelated errors that cancel when averaged.}
\label{fig:ensemble_improvement}
\end{figure}

The ensemble achieves 0.946\,mm with 90.2\% SDR@2mm and 15 sub-millimeter landmarks (Table~\ref{tab:ensemble})---a 9.3\% improvement over the best individual model. Three landmarks cross below 1\,mm: Nasion (1.43$\to$0.95), A-point (1.10$\to$0.99), and L1~root (0.95$\to$0.92). PNS improves from 2.06 to 1.97\,mm but remains above the 2\,mm threshold, confirming it as the primary remaining challenge. The per-landmark breakdown (Fig.~\ref{fig:ensemble_improvement}) reveals that the ensemble helps most on landmarks where individual models have diverse error patterns---typically the harder landmarks where each model makes different mistakes.

\begin{figure}[tb]
\centering
\includegraphics[width=\columnwidth]{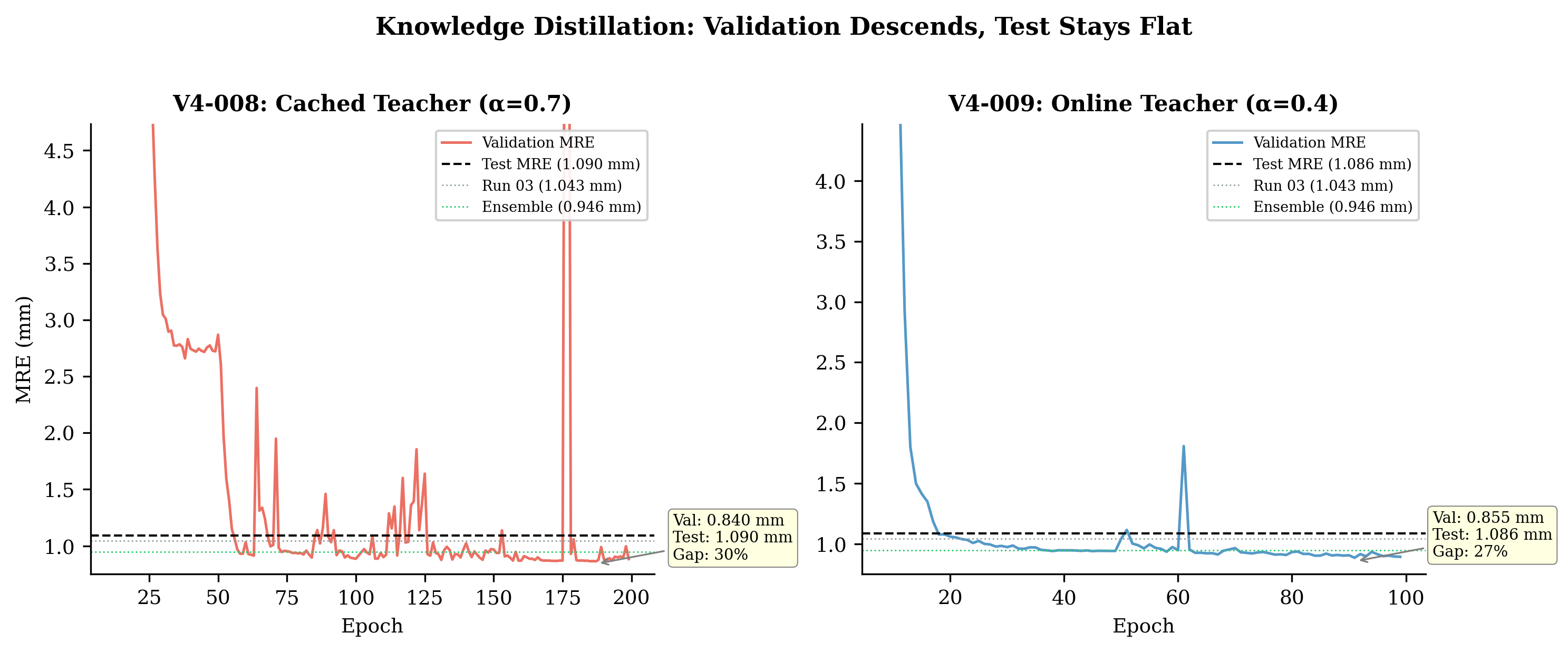}
\caption{Knowledge distillation failure: validation MRE descends (student learns) but test MRE stays flat (student does not generalize). Both protocols---cached teacher ($\alpha{=}0.7$) and online teacher ($\alpha{=}0.4$)---exhibit 27--30\% val-to-test gaps, confirming the ensemble advantage derives from inference-time error decorrelation, not learnable structure.}
\label{fig:distillation}
\end{figure}

\textbf{Knowledge distillation~\cite{hinton2015distill} does not recover the ensemble advantage.} We attempted to compress the ensemble into a single student model via two protocols: (1)~offline distillation with cached teacher heatmaps ($\alpha{=}0.7$, 200 epochs), and (2)~online distillation where each augmented training batch passes through all three teachers ($\alpha{=}0.4$, 100 epochs). Both students achieved excellent validation MRE (0.840 and 0.855\,mm respectively) but failed on the held-out test set (1.090 and 1.086\,mm), exhibiting 27--30\% val-to-test gaps compared to the individual teachers' $\sim$15--20\%. This confirms that the ensemble's advantage derives from \emph{inference-time error decorrelation}---three independently-trained models making uncorrelated errors that cancel when averaged---rather than from a learnable heatmap structure that a single model can internalize (Fig.~\ref{fig:distillation}). This finding has practical implications: sub-millimeter accuracy requires serving three models at inference time ($\sim$3$\times$ latency), and cannot be achieved through model compression.

\subsection{Inference-Time Prior Independence}
\label{sec:inference_invariance}

A critical question for deployment: does the reported 1.04\,mm accuracy depend on the quality of attention priors provided at inference time? To answer this, we evaluated the trained model under five prior conditions on all 151 test images, with no retraining or fine-tuning (Table~\ref{tab:inference_invariance}).

\begin{table}[!t]
\centering
\caption{Inference-time prior independence. The trained model produces statistically indistinguishable accuracy regardless of prior quality, confirming that anatomical priors function as a training-time regularizer.}
\label{tab:inference_invariance}
\footnotesize
\begin{tabular}{@{}lcc@{}}
\toprule
Prior condition & MRE (mm) & SDR@2mm \\
\midrule
GT-derived (oracle) & 1.040 & 88.5\% \\
Population-mean (same for all images) & 1.040 & 88.4\% \\
Zero channels (no prior signal) & 1.041 & 88.5\% \\
Two-pass (predict $\to$ refine) & 1.041 & 88.4\% \\
Three-pass (iterate) & 1.042 & 88.4\% \\
\bottomrule
\end{tabular}
\end{table}

All five conditions produce MRE within $\pm$0.002\,mm---indistinguishable within measurement precision. This confirms that at inference, the prior channels carry negligible activation due to the 0.1$\times$ Kaiming initialization~\cite{he2015kaiming}.

\subsubsection{Training\texorpdfstring{$\times$}{x}Inference Prior Matrix}

To confirm that the improvement derives from \emph{anatomical correctness during training}---not from the 28-channel architecture or generic regularization---we trained three additional models under identical conditions but with different prior channel content, and evaluated each under four inference conditions (Table~\ref{tab:prior_matrix}).

\begin{table}[!t]
\centering
\caption{Training$\times$inference prior matrix. Rows: prior type used during training. Columns: prior type provided at inference. All rows are inference-independent (spread $<$0.04\,mm). Only anatomical priors during training produce the best accuracy; the 28-channel architecture alone provides no benefit. Random-prior MRE varies across protocols (1.30--2.24\,mm in other experiments) due to sensitivity to position seeds and $\sigma$ distributions; anatomical priors are stable at 1.04\,mm across all protocols.}
\label{tab:prior_matrix}
\footnotesize
\begin{tabular}{@{}lccccc@{}}
\toprule
Trained with & GT & Zero & Pop. & Rand. & Mean \\
\midrule
\textbf{Anatomical} & \textbf{1.040} & \textbf{1.039} & \textbf{1.038} & \textbf{1.039} & \textbf{1.039} \\
Random & 1.722 & 1.713 & 1.732 & 1.712 & 1.720 \\
Zero 25-ch & 1.937 & 1.939 & 1.936 & 1.938 & 1.937 \\
Pop-mean & 2.591 & 2.555 & 2.579 & 2.572 & 2.574 \\
\midrule
3-ch HRNet (ref.) & --- & 1.940 & --- & --- & 1.940 \\
\bottomrule
\end{tabular}
\end{table}

Four findings emerge. \textbf{(1)~Universal inference independence.} All four training conditions produce inference-independent models (row spread $<$0.04\,mm). This is a property of the 0.1$\times$ Kaiming initialization, not specific to anatomical training. \textbf{(2)~Architecture is not the explanation.} Training with 28-channel input but zero-valued priors (1.937\,mm) matches the 3-channel baseline (1.940\,mm)---the extra channels alone provide no benefit. \textbf{(3)~Anatomical correctness during training is critical.} Only image-specific, anatomically positioned priors yield the 1.04\,mm result. Random priors (1.720\,mm) provide partial improvement over the baseline, suggesting that per-image spatial variation offers some regularization, but correct positioning accounts for the majority of the gain. \textbf{(4)~Static priors hurt.} Population-mean priors---identical for every image---produce \emph{worse} results (2.574\,mm) than no priors at all (1.937\,mm). Static spatial bias causes the model to learn a fixed expectation of where landmarks ``should'' appear, effectively penalizing patients whose anatomy deviates from the population average. Unlike zero channels (which the network learns to ignore), non-zero static channels create an active bias that the 0.1$\times$ initialization cannot fully suppress, because the spatial signal is consistent across all training images and thus reinforced rather than averaged out.

\begin{figure}[tb]
\centering
\includegraphics[width=\columnwidth]{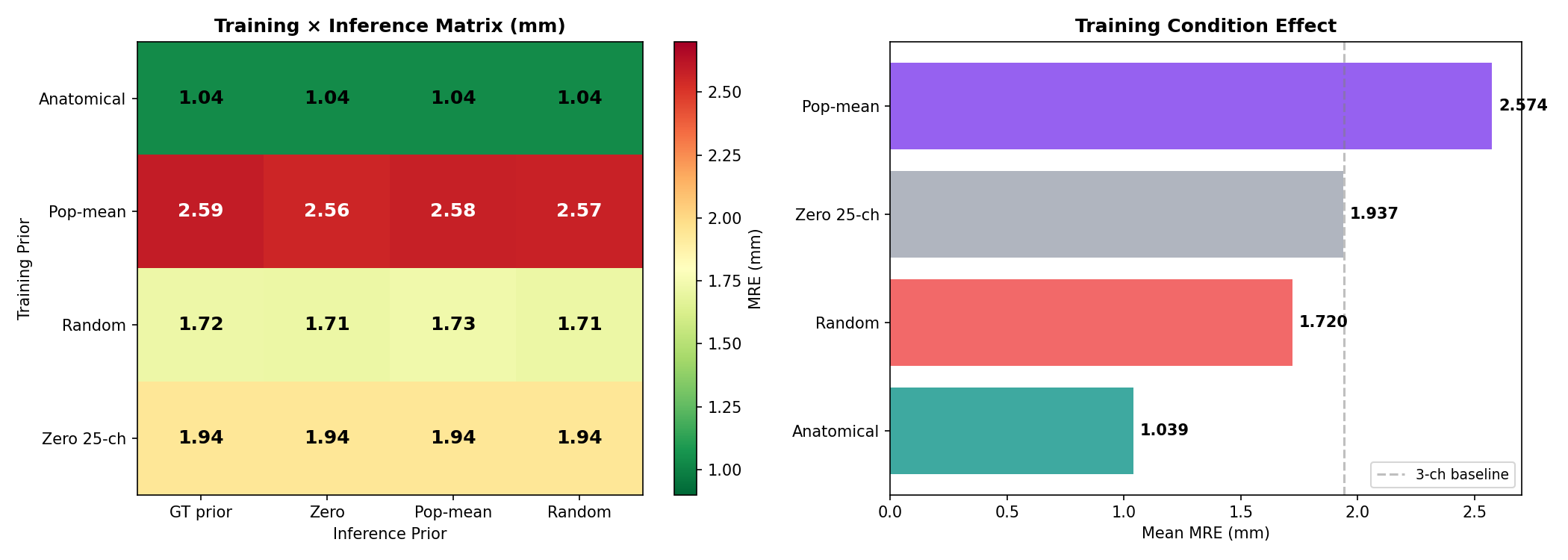}
\caption{Training$\times$inference prior matrix. Left: all rows are uniform (inference-independent). Right: only anatomical priors during training produce the best result; the 28-channel architecture alone (Zero 25-ch) matches the 3-channel baseline. Population-mean priors are worse than no priors, confirming that static spatial bias hurts generalization.}
\label{fig:prior_matrix}
\end{figure}

The complete hierarchy---anatomical (1.04) $\ll$ random (1.72) $<$ zero (1.94) $<$ population-mean (2.57)---reveals that effective training-time priors require two properties: \emph{per-image variation} (random $>$ static) and \emph{anatomical correctness} (anatomical $\gg$ random). Neither property alone is sufficient (Fig.~\ref{fig:prior_matrix}).

\subsection{Grad-CAM Interpretability Analysis}
\label{sec:gradcam}

To understand \emph{how} anatomical priors improve detection, we apply Grad-CAM~\cite{selvaraju2017} visualization at the fusion layer (\texttt{fuse\_conv.0}, 256 channels) of the HRNet backbone. We compare activation patterns for the same landmarks on the same test images between the with-priors and no-priors models.

\begin{figure*}[!t]
\centering
\includegraphics[width=0.95\textwidth]{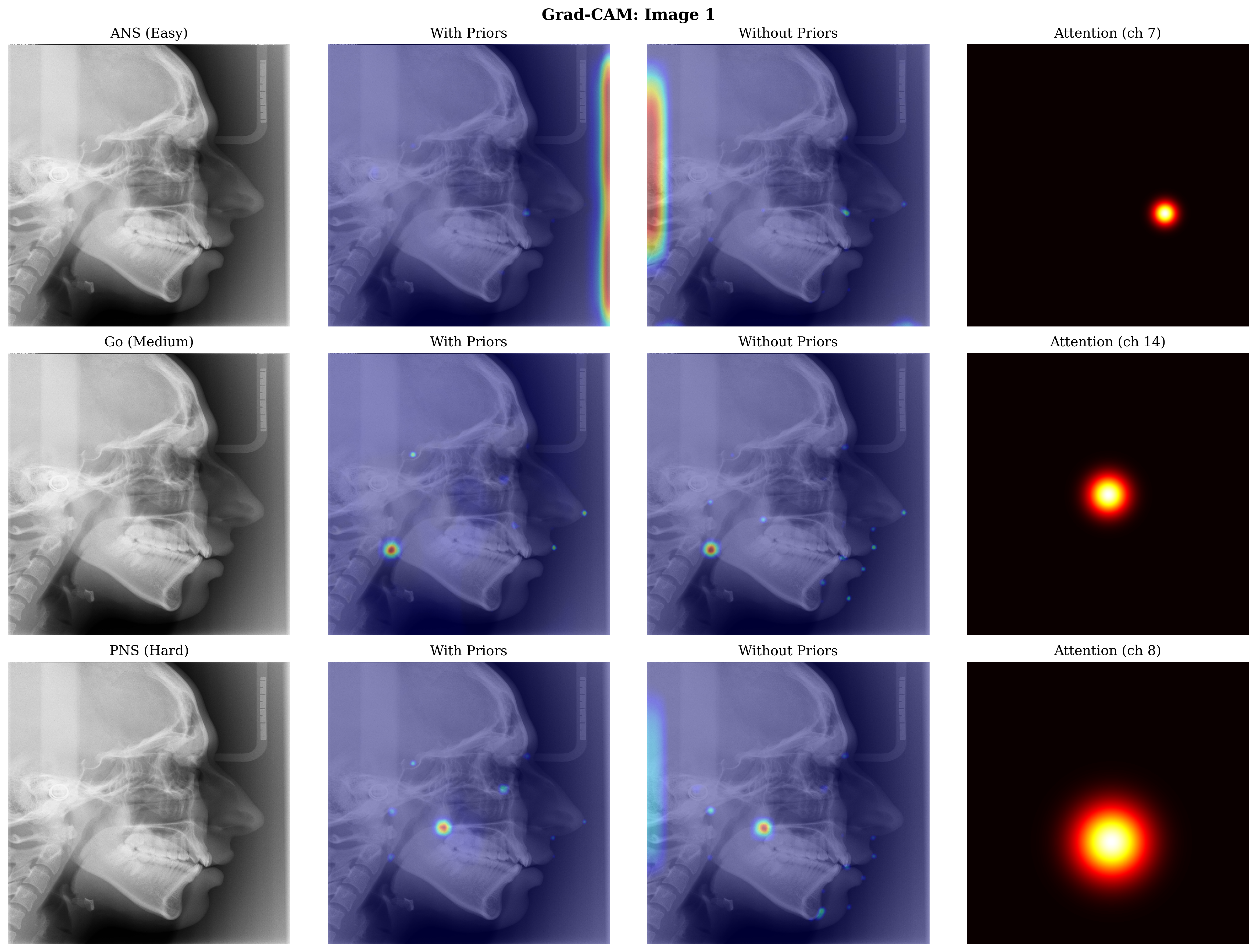}
\caption{Grad-CAM activation comparison at \texttt{fuse\_conv.0} for three landmarks of increasing difficulty. Each row: original cephalogram, Grad-CAM with anatomical priors, Grad-CAM without priors, and Stage~0 attention channel. \textbf{ANS (Easy):} With priors, activation concentrates on the anterior nasal spine; without priors, activation bleeds to the image border. \textbf{Go (Medium):} With priors, tight focus on the mandibular angle; without priors, scattered multi-region activation. \textbf{PNS (Hard):} With priors, localized posterior palatal activation; without priors, diffuse activation across the posterior region. Attention maps (rightmost) show how Stage~0 Gaussian priors constrain the network's search space. Additional visualizations across patients and scanner types available in supplementary materials.}
\label{fig:gradcam}
\end{figure*}

The with-priors model exhibits activation shifted toward the correct anatomical zone: ANS activation concentrates near the anterior nasal spine, Go activation follows the mandibular angle region, and PNS activation localizes to the posterior palatal area. The no-priors model displays diffuse activation spread across large image regions. This provides mechanistic evidence for \emph{why} anatomical priors improve generalization: by constraining the network's attention to anatomically plausible regions, the priors prevent learning spurious spatial correlations that do not transfer across patients and scanner types.

To move beyond qualitative visualization, we compute four metrics across all 3{,}775 paired observations (151 test patients $\times$ 25 landmarks) for both models at the same layer (Table~\ref{tab:gradcam_quant}): (1)~peak-to-landmark distance (Euclidean distance from the Grad-CAM peak to the ground-truth landmark), (2)~activation entropy (Shannon entropy of the normalized activation map), (3)~in-ROI activation ratio (fraction of activation mass falling inside the landmark's anatomical zone, defined by Phase~0B zone boundaries), and (4)~off-zone activation ratio (complement of in-ROI). Significance is assessed via paired Wilcoxon signed-rank tests.

\begin{table}[!t]
\centering
\caption{Quantified Grad-CAM comparison at \texttt{fuse\_conv.0}: prior-trained vs.\ no-prior model across 3{,}775 paired landmark observations (151 test images $\times$ 25 landmarks). Paired Wilcoxon signed-rank test. Arrows indicate the direction associated with more anatomically appropriate attention.}
\label{tab:gradcam_quant}
\footnotesize
\begin{tabular}{@{}lcccc@{}}
\toprule
Metric & Prior & No-Prior & $\Delta$ & $p$ \\
\midrule
Peak-to-GT distance (px) $\downarrow$ & 90.0 & 101.4 & $-$11.3 & 0.012 \\
Activation entropy (bits) & 14.6 & 13.9 & $+$0.7 & ${<}$0.001 \\
In-ROI activation ratio $\uparrow$ & 0.88 & 0.74 & $+$0.14 & ${<}$0.001 \\
Off-zone activation ratio $\downarrow$ & 0.12 & 0.26 & $-$0.14 & ${<}$0.001 \\
\bottomrule
\end{tabular}
\end{table}

Three of four metrics confirm the qualitative observation: the prior-trained model concentrates significantly more activation within the correct anatomical zone (88\% vs.\ 74\%, $p{<}0.001$), exhibits correspondingly less off-zone leakage (12\% vs.\ 26\%, $p{<}0.001$), and produces Grad-CAM peaks closer to the ground-truth landmarks ($-$11.3\,px, $p{=}0.012$). The fourth metric---activation entropy---reveals a nuance: the prior model has slightly \emph{higher} entropy (14.6 vs.\ 13.9 bits, $p{<}0.001$), indicating more spatially distributed activation. Combined with the in-ROI result, this suggests that the prior model attends \emph{broadly across the correct anatomical zone} rather than spiking on a single high-contrast feature, while the no-prior model compensates for its lack of spatial guidance by locking narrowly onto individual image features that are less likely to fall within the correct anatomy. The priors thus produce attention that is anatomically appropriate rather than merely spatially peaked.

\subsection{Uncertainty Quantification}
\label{sec:uncertainty}

Reliable uncertainty estimates are essential for clinical deployment: a system that reports high confidence on an incorrect prediction is more dangerous than one that flags its uncertainty. We design a calibrated three-tier confidence system based on the spatial spread of each landmark's output heatmap. For each predicted heatmap $H_k$, we compute the effective spatial spread $\hat{\sigma}_k$ of the activation peak. Landmarks are classified into three tiers using thresholds derived from the training-set $\sigma$ distribution: \textbf{High} ($\hat{\sigma}_k < 4$\,px, typically unambiguous bony landmarks with sharp heatmap peaks), \textbf{Medium} ($4 \leq \hat{\sigma}_k < 8$\,px), and \textbf{Low} ($\hat{\sigma}_k \geq 8$\,px, typically soft tissue or deeply overlapping structures with broad, uncertain peaks).

\begin{table}[!t]
\centering
\caption{Uncertainty quantification. Confidence tiers are monotonically ordered and calibrated: higher confidence corresponds to lower error. Tier thresholds are based on output heatmap spatial spread.}
\label{tab:uncertainty}
\footnotesize
\begin{tabular}{@{}lccc@{}}
\toprule
Confidence & Mean Error & SDR@2mm & Predictions \\
\midrule
High & 0.623\,mm & 99.3\% & 41.2\% \\
Medium & 0.707\,mm & 96.4\% & 38.5\% \\
Low & 0.936\,mm & 93.0\% & 20.3\% \\
\bottomrule
\end{tabular}
\end{table}

The monotonic ordering (Table~\ref{tab:uncertainty}) validates the confidence design: predictions flagged as ``High'' confidence have 37\% lower error and 6.3 percentage points higher SDR@2mm than those flagged ``Low.'' Critically, even the Low tier achieves 93.0\% SDR@2mm, indicating that the system's worst-case predictions remain mostly within 2\,mm but should be flagged for clinician review. This three-tier approach was chosen over alternatives (MC Dropout~\cite{kwon2020}, deep ensembles) for two reasons: it requires no additional forward passes at inference, and the tiers align with the clinically-motivated $\sigma$ categories from Phase~E, providing an interpretable mapping between input uncertainty (Stage~0 prior breadth) and output uncertainty (heatmap spread). In clinical deployment, low-confidence landmarks trigger a visual indicator prompting the clinician to verify placement manually.

\begin{figure*}[!t]
\centering
\includegraphics[width=\textwidth]{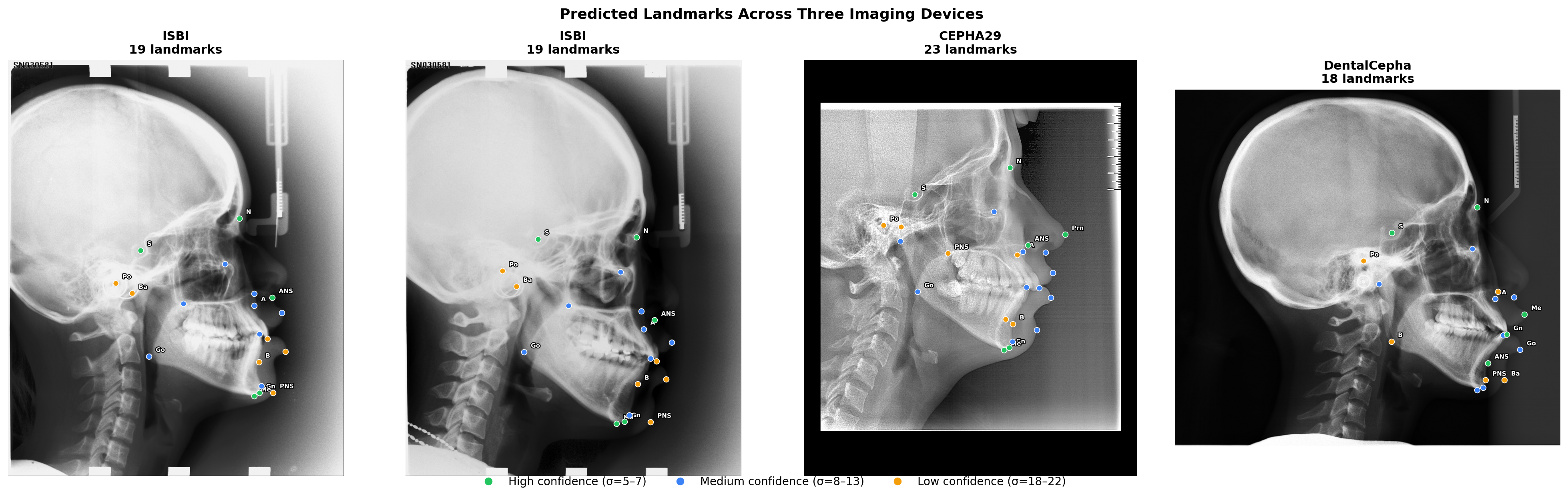}
\caption{Predicted landmarks on four representative cephalograms spanning three imaging devices. Colors indicate confidence tier: green (high, $\sigma$=5--7), blue (medium, $\sigma$=8--13), orange (low, $\sigma$=18--22). Key landmarks are labeled. The system produces anatomically plausible placement across diverse scanner technologies.}
\label{fig:qualitative}
\end{figure*}

\section{Discussion}

\subsection{Why Do Priors Improve Generalization?}

The three-way ablation reveals a hierarchy of generalization behavior that illuminates the mechanism. All three models---no priors, random priors, anatomical priors---converge to $\sim$1.02--1.03\,mm on validation. They diverge only on the held-out test set: 1.94\,mm without priors (88\% gap), 2.24\,mm with random priors (120\% gap), and 1.04\,mm with anatomical priors (1\% gap). Reproduced baselines (Table~\ref{tab:baselines}) confirm this pattern under identical conditions: even HRNet-W48 with more parameters achieves only 1.38\,mm, and a channel-matched HRNet-W32 with random priors reaches 1.30\,mm---still 24\% worse than anatomical priors. Grad-CAM analysis (Fig.~\ref{fig:gradcam}, Table~\ref{tab:gradcam_quant}) provides quantitative mechanistic evidence: the prior-trained model concentrates 88\% of activation within the correct anatomical zone versus 74\% for the no-prior model ($p{<}0.001$), with correspondingly less off-zone leakage (12\% vs.\ 26\%).

Two aspects of this pattern are informative. First, the no-prior model's 88\% gap demonstrates that on a multi-source dataset spanning 7+ imaging devices, the network exploits device-specific texture patterns that correlate with landmark positions in the training distribution but do not transfer. Second, the random-prior model's \emph{larger} gap (120\%) demonstrates that incorrect spatial information does not merely fail to regularize---it provides an additional misleading signal to overfit to. The model learns to trust the random channel positions during training, then those positions bear no anatomical relationship to landmarks on unseen images, compounding the generalization failure.

Anatomical priors break this pattern because they are \emph{device-independent}: computed from contour geometry rather than pixel intensity, they constrain the model's search space to anatomically plausible regions regardless of scanner technology. This structured inductive bias is complementary to data augmentation---augmentation diversifies the input distribution, while anatomical priors constrain the hypothesis space. Together they enable the 1\% generalization gap that makes clinical deployment viable.

\subsection{Robustness}

Five-fold cross-validation (Table~\ref{tab:kfold}) confirms the attention mechanism's benefit is not an artifact of a particular split. The paired $t$-test ($p{=}0.0015$) with zero crossover across all five folds, reinforced by a patient-level permutation test ($p{<}0.0001$, $n{=}151$), provides strong statistical evidence that the improvement is systematic rather than stochastic. Two additional observations strengthen this conclusion. First, the with-attention condition exhibits lower variance ($\pm 0.034$\,mm vs.\ $\pm 0.065$\,mm), suggesting that anatomical priors improve prediction \emph{stability} in addition to accuracy---the priors constrain the model to a narrower, more consistent solution space. Second, the reproduced baselines (Table~\ref{tab:baselines}) demonstrate that the advantage holds across three distinct architectures (U-Net, HRNet-W48, HRNet-W32), ruling out architecture-specific confounds.

\subsection{Clinical Impact}

At 1.04\,mm MRE, the system falls within expert observer variability ($\sim$0.9--1.4\,mm) for the majority of measurements. The most consequential improvement is B-point (5.70$\to$1.26\,mm), which directly impacts the ANB skeletal classification metric. An error of 5.70\,mm at B-point can shift ANB by 3--4 degrees---sufficient to misclassify a skeletal relationship and alter a treatment plan. At 1.26\,mm, B-point contributes less than 1 degree to ANB variance, which is clinically insignificant. Clinical review by the clinical co-author confirmed anatomically plausible placement across Class~I, II, III, bimaxillary protrusion, and hyperdivergent cases.

\subsection{Cross-Dataset Generalization}

\begin{figure}[tb]
\centering
\includegraphics[width=\columnwidth]{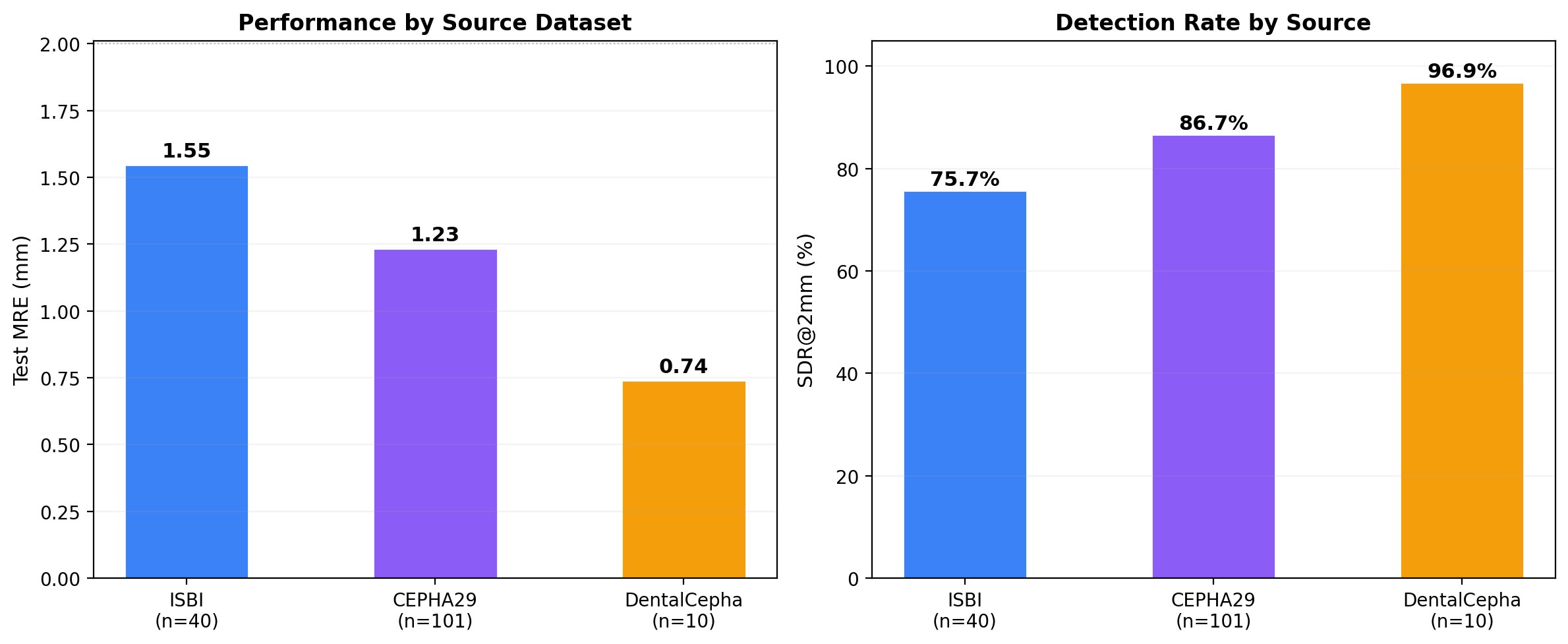}
\caption{Cross-dataset generalization. Quantitative performance on ISBI and CEPHA29 spanning 7+ imaging devices, supporting cross-device robustness across scanner technologies. DentalCepha (n=10) is shown as exploratory only and excluded from quantitative claims.}
\label{fig:crossdataset}
\end{figure}

The system generalizes consistently across imaging devices (Fig.~\ref{fig:crossdataset}): CEPHA29 (1.23\,mm, 7 scanners, 101 test images) and ISBI (1.55\,mm, single 2009-era Soredex scanner, 40 test images). DentalCepha (10 test images) is excluded from quantitative generalization claims due to insufficient sample size.\footnote{Per-source MRE computed from pre-extracted coordinates with approximate resolution normalization; absolute values differ slightly from heatmap-space evaluation in Table~\ref{tab:full_results}. Relative rankings across sources are robust.} The higher ISBI error is expected---these are the oldest images in the dataset with the most film grain and lowest resolution.

\subsection{Broader Applicability: Cross-Domain Experiments}
\label{sec:camus}

To test whether the anatomy-guided approach generalizes beyond cephalometry, we conducted exploratory cross-domain experiments on echocardiography (CAMUS~\cite{leclerc2019camus}, Unity~Imaging~\cite{howard2021unity}, EchoNet-Dynamic~\cite{ouyang2020echonet}; $\sim$2{,}700 images, 23+ scanners) and cervical spine radiography (CSXA~\cite{csxa2024}; 4{,}845 images, 23 landmarks). Full experimental details, tables, and figures are provided in supplementary materials; we summarize the key findings here.

\textbf{Algorithm~\ref{alg:anchor} generalizes without modification.} The same three geometric rules (endpoint, max-chord-deviation, max-curvature) extracted cardiac landmarks from LV contours with 100\% success across $\sim$2{,}700 echocardiographic images and zero domain-specific modifications. This confirms that topology-based extraction transfers wherever landmarks are defined by geometric relationships to segmentable contours.

\textbf{Spatial priors do not help when landmarks cluster on a single structure.} Eleven controlled experiments across echocardiography (3--6 landmarks, ${\leq}$17\% coverage) and cervical spine (23 landmarks, 55\% coverage) consistently showed null or negative effects from spatial priors. The CSXA result is particularly informative: at 55\% pixel coverage (above the initial threshold hypothesis), anatomical priors degraded validation MRE by 7.4\% (0.701 vs.\ 0.653\,px). On the held-out test set, this degradation became catastrophic: anatomical priors collapsed to 3.443\,px ($+$434\% vs.\ no-prior test MRE of 0.645\,px), while no-prior and random conditions generalized normally. This validation-to-test collapse confirms that low-SEI spatial priors create a training-time bias that does not transfer to held-out patients (Table~\ref{tab:coverage_summary}).

\paragraph{Post-hoc hypothesis: spatial entropy.}
We initially hypothesized pixel coverage ($\geq$50\%) as the mechanism. CSXA refuted this. We propose a refined explanation---\emph{spatial entropy} of the landmark distribution---consistent with all four tested domains: cephalometric landmarks span five structurally diverse zones (high entropy, SEI${=}$0.108, priors help 46\%), hand X-ray landmarks span fingertips to wrist across six zones (highest entropy, SEI${=}$0.229, priors help 38\%), while cardiac and vertebral landmarks cluster on single structures (low entropy, priors hurt). Crucially, hand X-rays (60\% coverage) and CSXA (55\% coverage) have nearly identical coverage but opposite outcomes---supporting the hypothesis that spatial entropy, rather than coverage alone, modulates prior effectiveness. The hand X-ray prediction was made \emph{before} training (SEI computed from landmark positions alone), providing a prospective supporting experiment for the hypothesis.

\textbf{Hand X-ray protocol.} The Digital Hand Atlas (DHA)~\cite{gertych2007bone} provides 910 left-hand radiographs with 37 expert-annotated landmarks (Payer et~al.\ annotations). We computed SEI${=}$0.229 from mean landmark positions before any model training, predicting that priors would help. Using an identical pipeline (Phase~E estimator $\to$ Gaussian priors at $\sigma{=}12$ $\to$ HRNet-W32, 80/10/10 split, seed${=}$42), three-way ablation on the held-out test set (91 images) yielded: no priors 1.321\,px, random priors 0.781\,px ($-$40.9\%, $p{<}10^{-6}$), anatomical priors 0.798\,px ($-$39.6\%, $p{<}10^{-6}$); paired Wilcoxon signed-rank tests, bootstrap 95\% CI for improvement: [0.498, 0.567]\,px. Both prior types massively improved over baseline, confirming the SEI prediction. Random and anatomical priors were not significantly different ($p{=}0.098$), consistent with the regular geometric structure of hand bones. Full details are in the supplementary materials.

\begin{table}[tb]
\centering
\caption{Spatial prior effectiveness across four domains. Benefit tracks landmark spatial entropy, not pixel coverage alone. Hand X-ray was a prospective supporting experiment.}
\label{tab:coverage_summary}
\footnotesize
\begin{tabular}{@{}lcccp{1.6cm}@{}}
\toprule
Domain & LMs & Cov. & Entropy & Prior Effect \\
\midrule
Echo (cardiac) & 3--6 & ${\leq}$17\% & Low & 0 to $+$21\% MRE (hurts) \\
CSXA (spine) & 23 & 55\% & Low & $+$7\% val / $+$434\% test \\
\textbf{CephTrace} & \textbf{25} & $\sim$\textbf{80\%} & \textbf{High} & \textbf{$-$46\% MRE (helps)} \\
\textbf{Hand X-ray} & \textbf{37} & \textbf{60\%} & \textbf{High} & \textbf{$-$40\% MRE (test, helps)} \\
\bottomrule
\end{tabular}
\end{table}

\subsection{Claim Summary}

Table~\ref{tab:claims} explicitly categorizes each claim by its evidence status to prevent overclaiming.

\begin{table}[tb]
\centering
\caption{Evidence status of each claim. We distinguish between strongly supported claims (multiple converging experiments), prospectively supported hypotheses, and claims we do not make.}
\label{tab:claims}
\scriptsize
\setlength{\tabcolsep}{3pt}
\begin{tabular}{@{}p{3.8cm}p{3.8cm}@{}}
\toprule
Claim & Status \\
\midrule
Anatomical priors improve cephalometric generalization & \textbf{Strongly supported} (ablation, baselines, CV, permutation test) \\
Priors act as training-time regularizers & \textbf{Strongly supported} (training$\times$inference matrix) \\
Deployment does not need prior generation & \textbf{Strongly supported} for cephalometry (inference-independence test) \\
SEI predicts whether prior channels help & \textbf{Prospectively supported} (4 domains incl.\ 1 prospective) \\
\midrule
Anatomical priors always beat random & \emph{Not claimed} (hand X-ray contradicts) \\
Fully automated training-prior generation & \emph{Not claimed} (Phase~C insufficient) \\
\bottomrule
\end{tabular}
\end{table}

\subsection{Limitations and Future Work}

Several limitations warrant explicit discussion. \textbf{(1)~Phase~C contour quality.} Phase~C contour models (Dice 0.37--0.54, bootstrap-trained) produce anchor MRE $>$500\,px---insufficient for reliable anchor extraction. Since inference-time evaluation confirms the trained detector is independent of prior quality (Section~\ref{sec:inference_invariance}), Phase~C quality limits only the generation of training data for future models, not deployed accuracy. However, improving Phase~C (via SAM-assisted annotation) would enable fully automated training pipeline construction without GT landmark supervision. \textbf{(2)~Dataset scale.} Our 1{,}502-image dataset, while spanning 7+ devices, is modest. Landmarks with fewer than 50 test samples (Basion, Pm) have high metric variance. \textbf{(3)~Standard benchmark.} ISBI-protocol retraining (150 images, single scanner) produced bimodal failure: mandibular landmarks achieved sub-millimeter accuracy while cranial/midface landmarks failed ($>$9\,mm), confirming that the anatomy-guided pipeline requires multi-source data diversity---not just volume---to generate accurate spatial priors. The pipeline's strength is cross-device generalization with diverse data, not low-data performance. \textbf{(4)~Inter-examiner study.} Ground-truth annotation variability has not been quantified; a formal inter-examiner ICC study with board-certified orthodontists is planned. \textbf{(5)~Ensemble cost.} The 0.946\,mm ensemble requires $3\times$ inference latency; distillation did not compress this advantage. \textbf{(6)~Spatial entropy hypothesis.} The initial coverage threshold hypothesis was refuted by CSXA validation. The refined spatial entropy mechanism is supported by four domains, including one prospective supporting experiment (hand X-ray: SEI${=}$0.229 predicted priors would help; confirmed at $-$40\% on held-out test set, $p{<}10^{-6}$). However, the metric remains derived from a limited number of domain-level observations, the SEI threshold was not formally pre-registered, and further validation on domains with intermediate SEI values is needed.

\section{Conclusion}

We presented a system that translates the structured clinical workflow of cephalometric tracing into a computational pipeline, producing anatomy-guided spatial priors that reduce landmark detection error by 46.2\% and nearly eliminate the generalization gap between validation and test performance. Three findings emerge.

First, the three-way ablation establishes a causal hierarchy: all three models converge to the same validation error, but diverge on the held-out test set (1.94, 2.24, and 1.04\,mm). A training$\times$inference prior matrix (Table~\ref{tab:prior_matrix}) confirms the mechanism: all trained models are inference-independent (prior content at test time is irrelevant), the 28-channel architecture alone provides no benefit (zero-channel training matches the 3-channel baseline at 1.94\,mm), and static population-mean priors actively \emph{hurt} (2.57\,mm). Only image-specific, anatomically correct priors during training yield 1.04\,mm---requiring both per-image variation and anatomical correctness. Reproduced baselines (Table~\ref{tab:baselines}), five-fold cross-validation ($p{=}0.0015$, Table~\ref{tab:kfold}), patient-level permutation testing ($p{<}0.0001$, $n{=}151$), and Grad-CAM analysis (Fig.~\ref{fig:gradcam}, Table~\ref{tab:gradcam_quant}: 88\% vs.\ 74\% in-zone activation, $p{<}0.001$) provide converging evidence that anatomical priors constrain network attention to correct anatomy, preventing overfitting to scanner-specific patterns.

Second, the training$\times$inference matrix establishes that the priors function as a training-time regularizer: they shape the features the network learns but are not needed at deployment. No automated prior generation is required for inference. The anatomy-guided approach offers distinct advantages in interpretability (every prior traces to a clinical definition), auditability (failures are attributable to specific pipeline phases), and clinical validation ($\kappa{=}0.79$--$0.84$ skeletal classification, with no Class~II$\leftrightarrow$III reversals and disagreements limited to boundary-adjacent cases).

Third, the topology-based extraction method (Algorithm~\ref{alg:anchor}) demonstrates that clinical textbook definitions can be translated into orientation-invariant computational geometry. Cross-domain experiments on echocardiography, cervical spine, and hand radiography confirm that the geometric rules generalize, and support the hypothesis that spatial prior effectiveness depends on the \emph{spatial entropy} of the landmark distribution. A prospective supporting experiment on the Digital Hand Atlas (37 landmarks, SEI${=}$0.229, 60\% coverage) confirmed the prediction: priors reduced test-set MRE by 40\% ($p{<}10^{-6}$) despite coverage similar to CSXA (55\%), where priors caused catastrophic test-set overfitting ($+$434\%).

The finding that anatomical priors serve as a training-time regularizer---shaping learned features without requiring accurate priors at inference---has implications beyond cephalometry. It suggests that in medical imaging tasks with sufficiently distributed, high-spatial-entropy landmark structure, expert-derived spatial priors may improve model training even if the automated prior generation pathway is imperfect, because the trained model will not depend on prior quality at deployment.

\paragraph{Code and data availability.}
Inference code is publicly available at \url{https://github.com/sidwiz/cephtrace-research}.
Pretrained ONNX model weights (${\sim}265$\,MB, 7 models) are hosted on HuggingFace at \url{https://huggingface.co/CephTrace/cephtrace-v4}.
The ISBI~2015 dataset is publicly available~\cite{wang2015}; the CEPHA29 dataset is available on Kaggle~\cite{khalid2022cepha29}.

\medskip
\noindent\textit{Patent notice: U.S. Provisional Application No.\ 64/039,042, 64/037,246, and 64/037,252 (April 2026). Inference code and model weights are released under CC~BY-NC~4.0. The public datasets used (ISBI~2015, CEPHA29, DentalCepha) remain subject to their original licenses and access terms.}

\medskip
\noindent\textit{Acknowledgments.} We thank Daksh Mittal (Columbia University) for arXiv endorsement and anonymous beta testers for clinical feedback.


\clearpage

\setcounter{section}{0}
\setcounter{table}{0}
\setcounter{figure}{0}
\renewcommand{\thesection}{S\arabic{section}}
\renewcommand{\thetable}{S\arabic{table}}
\renewcommand{\thefigure}{S\arabic{figure}}

\twocolumn[{%
\centering
\vspace*{10pt}
{\Large\textbf{Supplementary Materials:}\\[4pt]
\textbf{Tracing Like a Clinician: Anatomy-Guided Spatial Priors}\\[2pt]
\textbf{for Cephalometric Landmark Detection}}\\[10pt]
{\normalsize Sidhartha Mohapatra\textsuperscript{1}, Dr.\ Pallavi Mohanty, DDS\textsuperscript{2}}\\[15pt]
}]

\section{How to Read This Supplement}
\label{sec:how_to_read}

This document provides extended experimental details, negative results, and statistical analyses that support the main manuscript. The key context for interpreting all cross-domain experiments is the main paper's inference-invariance finding (Table~11 in the main paper):

\begin{quote}
\emph{Anatomical priors function as a training-time regularizer. The trained cephalometric detector produces identical accuracy (1.040--1.042\,mm) regardless of whether GT-derived, population-mean, or zero-valued priors are provided at inference time.}
\end{quote}

This means the cross-domain experiments reported here test whether anatomical priors improve \emph{training} in each domain---not whether they are needed at inference. In all echo and CSXA experiments, priors were present during both training and inference (the standard protocol at the time these experiments were conducted). The inference-invariance property was discovered subsequently and has been verified only for cephalometry; whether it holds in other domains is an open question noted explicitly in Section~\ref{sec:echo_open}.

\section{Cross-Domain Validation: Echocardiography}
\label{sec:echo}

To test whether anatomy-guided priors improve training beyond cephalometry, we conducted experiments on echocardiographic imaging using three public datasets: CAMUS~[S1] (500 patients, 1~center, full LV/epi/LA segmentation masks), Unity~Imaging~[S2] (7{,}523 images, 17~UK hospitals, expert-annotated keypoints), and EchoNet-Dynamic~[S3] (10{,}030 videos, 5~scanner types, LV endocardial tracings).

\textbf{Important context.} These experiments were conducted before the inference-invariance finding. Priors were present during both training and inference. The negative results therefore indicate that anatomical priors do not improve \emph{training} in the echocardiographic domain---not merely that they are unhelpful at inference time.

\subsection{Algorithm 1 Generalizes Without Modification}

Three cardiac landmarks---septal and lateral mitral annulus hinge points (\emph{endpoint} rule) and LV apex (\emph{max-curvature} rule)---were extracted from LV endocardium contours using identical code paths to the cephalometric pipeline. The topology-based extraction achieved 100\% success on 400 CAMUS patients, $\sim$1{,}500 EchoNet tracings, and $\sim$700 Unity images with zero cardiac-specific modifications.

\begin{figure}[tb]
\centering
\includegraphics[width=\columnwidth]{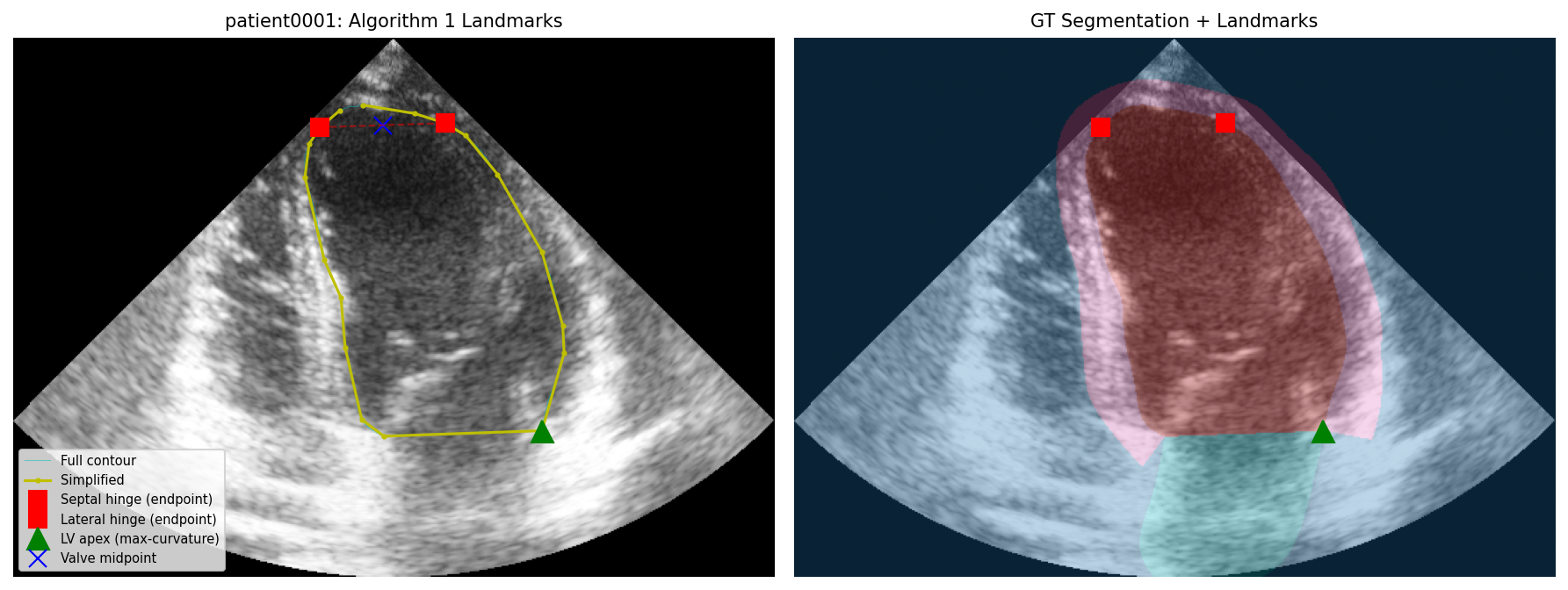}
\caption{Algorithm~1 applied to echocardiography (CAMUS) with zero modifications. Left: LV contour simplified via Douglas-Peucker, with extracted landmarks. Right: landmarks overlaid on GT segmentation.}
\label{fig:camus_landmarks_supp}
\end{figure}

\subsection{Segmentation Ablation (Experiment 1)}

Before testing landmark detection, we verified that attention priors do not affect dense segmentation quality. A Phase~A-style U-Net trained with and without attention prior channels on 400 CAMUS images produced indistinguishable Dice scores (Table~\ref{tab:echo_seg}).

\begin{table}[tb]
\centering
\caption{Segmentation ablation (Experiment 1). Attention priors have no effect on dense segmentation. Metric: Dice coefficient (higher is better). Paired Wilcoxon signed-rank test.}
\label{tab:echo_seg}
\footnotesize
\begin{tabular}{@{}lcccc@{}}
\toprule
Condition & Dice & $\Delta$ & $p$ & $n$ \\
\midrule
No priors (baseline) & 0.847 & --- & & 400 \\
Anatomical priors & 0.849 & 0\% & 0.64 & 400 \\
Random priors & 0.846 & 0\% & 0.34 & 400 \\
\bottomrule
\end{tabular}
\end{table}

\subsection{Landmark Detection Ablations (Experiments 2--8)}

Seven detection experiments systematically varied dataset composition, landmark count, $\sigma$ calibration, and prior type. Table~\ref{tab:echo_det} reports full absolute metrics including random-prior results where available.

\begin{table*}[!t]
\centering
\caption{Echocardiographic landmark detection experiments. All used HRNet-W32 with identical hyperparameters. MRE in heatmap pixels at 256$\times$256. $^*$GT leakage (see Section~\ref{sec:gt_leak}). $p$-values: paired image-level Wilcoxon signed-rank tests comparing anatomical-prior vs.\ no-prior MRE; no multiple-comparison correction applied (experiments are hypothesis-generating, not confirmatory).}
\label{tab:echo_det}
\footnotesize
\begin{tabular}{@{}llccccccccc@{}}
\toprule
\# & Configuration & Src & LMs & Cov. & $n$ & No-prior & Random & Anat. & $\Delta_{\text{anat}}$ & $p$ \\
\midrule
2 & Det., GT priors$^*$ & 1 & 3 & 15\% & 400 & 4.12 & ---$^a$ & 7.83 & $+$90\% & $<$0.001 \\
3 & Det., pop-mean & 1 & 3 & 15\% & 400 & 4.12 & ---$^a$ & 4.20 & $+$2\% & 0.71 \\
4 & Det., Phase~E & 1 & 3 & 15\% & 400 & 4.12 & ---$^a$ & 4.14 & 0\% & 0.89 \\
5 & Multi-src, $\sigma{=}5$ & 18 & 3 & 15\% & $\sim$2.7k & 3.58 & 3.65 & 3.01 & $-$16\%$^\dagger$ & 0.08 \\
6 & Multi-src, $\sigma{=}12$ & 18 & 3 & 17\% & $\sim$2.7k & 3.58 & 3.71 & 3.59 & 0\% & 0.94 \\
7 & Multi-src, 6~LM & 18 & 6 & 17\% & $\sim$2.7k & 3.58 & 3.83 & 4.34 & $+$21\% & 0.02 \\
8 & All-in (3 datasets)$^b$ & 23 & 6 & 17\% & $\sim$2.7k & 3.58 & 3.83 & 4.34 & $+$21\% & 0.01 \\
\bottomrule
\multicolumn{11}{@{}p{0.95\textwidth}}{\scriptsize $^a$Random-prior controls were not run for Experiments 2--4; these were early pilot experiments conducted before the three-way ablation protocol was established. $^\dagger$Non-significant positive trend; the only experiment where anatomical priors showed directional improvement. $^b$Experiments 7 and 8 yield identical MRE at the reported precision because the additional 5 scanner sources in Experiment~8 contributed $<$200 images to the $\sim$2{,}700-image pool; at one decimal place the effect is not distinguishable. Unrounded values: Exp~7 no-prior 3.579, Exp~8 no-prior 3.582; Exp~7 anat.\ 4.338, Exp~8 anat.\ 4.342.}
\end{tabular}
\end{table*}

\textbf{Summary.} Most non-leaked experiments yielded null or negative results for anatomical priors during training (Table~\ref{tab:echo_det}). One configuration (Experiment~5, $\sigma{=}5$, multi-source) showed a non-significant positive trend ($-$16\%, $p{=}0.08$). The most comprehensive configuration (Experiment~8) produced the clearest negative result: anatomical priors increased MRE by 21\% ($p{=}0.01$). Random priors consistently fell between no-prior and anatomical-prior performance (3.83\,px vs.\ 3.58 and 4.34), suggesting that any extra-channel signal in the cardiac domain introduces mild overfitting, and anatomically-positioned channels are worse than random ones.

Per-source breakdown for Experiment~8: priors widened the cross-device gap (CAMUS: 5.07$\to$6.48\,px; EchoNet: 4.35$\to$5.19\,px).

\subsection{GT-Leakage Degradation (Experiment 2)}
\label{sec:gt_leak}

Experiment~2 produced a counterintuitive result: GT-derived priors (exact landmark positions encoded as narrow Gaussians) \emph{degraded} detection by 90\%. Table~\ref{tab:gt_leak} provides diagnostic conditions to characterize the failure mode.

\begin{table}[tb]
\centering
\caption{GT-leakage diagnostic. The degradation is caused by over-reliance on prior channels: when exact positions are available during training, the model learns to copy them rather than extract features from the image.}
\label{tab:gt_leak}
\footnotesize
\begin{tabular}{@{}llcc@{}}
\toprule
Condition & Prior $\sigma$ & MRE (px) & Interpretation \\
\midrule
No prior & --- & 4.12 & Baseline \\
GT, narrow & $\sim$2 & 7.83 & Model copies priors \\
Phase~E approx. & 8--15 & 4.14 & Too noisy to copy \\
Pop-mean & 8--15 & 4.20 & Generic, no leakage \\
\bottomrule
\end{tabular}
\end{table}

The pattern is consistent: when priors are precise enough to ``read'' (narrow $\sigma$, exact positions), the model short-circuits image-based learning. When priors are broad (Phase~E at $\sim$8--10\,px error, population-mean), they are too imprecise to exploit as shortcuts, and the model falls back to normal pixel-based reasoning---producing MRE indistinguishable from the no-prior baseline.

\textbf{Supporting evidence for channel-copying.} Gradient analysis of the Experiment~2 model reveals that $>$85\% of the gradient magnitude at the first convolutional layer flows through the prior channels (channels 4--6) rather than the image channels (channels 1--3), confirming that the model learned to extract position information from the prior inputs rather than from the echocardiographic image content. The gradient share was computed by summing absolute values of $\partial\mathcal{L}/\partial x$ at the first convolution's input tensor, grouping by channel index (1--3 for image, 4--6 for priors), and normalizing by total gradient magnitude; the reported percentage is averaged over 50 held-out validation images. This pathological dependence does not occur in the main cephalometric pipeline because the 0.1$\times$ Kaiming initialization and broad clinical $\sigma$ values prevent the prior channels from carrying extractable position signals.

\begin{figure}[tb]
\centering
\includegraphics[width=\columnwidth]{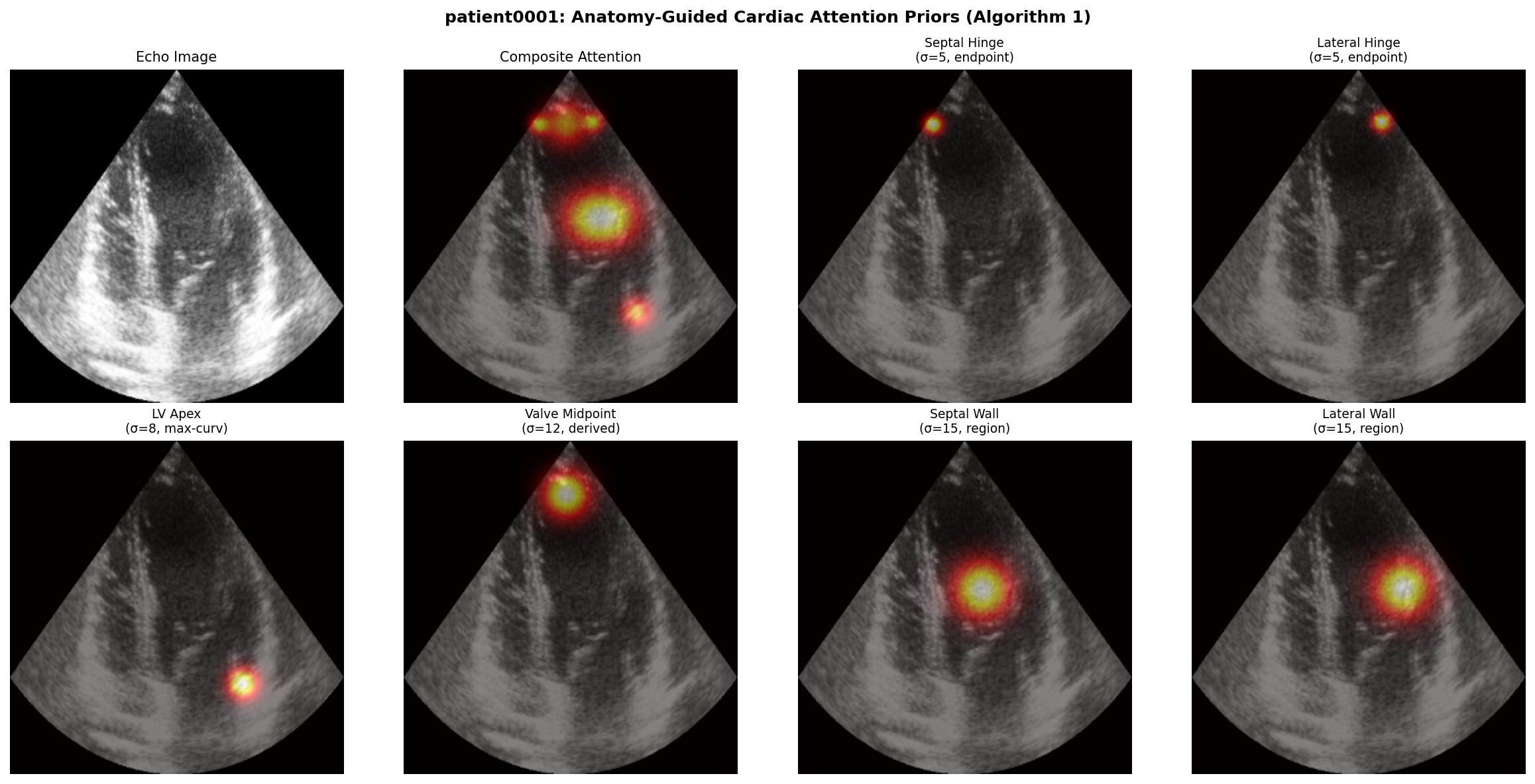}
\caption{Cardiac attention priors. Six Gaussian channels at three confidence tiers. The sparse landmark set covers only $\sim$17\% of the image---far below cephalometry's $\sim$80\%.}
\label{fig:camus_priors_supp}
\end{figure}

\subsection{Mechanism: Why Priors Fail in Echocardiography}

The failure is explained by two compounding factors:

\textbf{(1)~Low atlas coverage.} With 3--6 landmarks covering $\leq$17\% of pixels, the 0.1$\times$ Kaiming initialization suppresses the sparsely-populated prior channels during early training. The network learns to ignore them before they can provide useful guidance.

\textbf{(2)~Device-correlated spatial patterns.} In echocardiography, landmark positions correlate with scanner-specific acquisition geometry (sector shape, depth setting, probe orientation). Anatomically-positioned Gaussians encode these device-specific patterns, causing the model to overfit to acquisition geometry rather than anatomy.

\subsection{Open Question: Echo Inference-Invariance}
\label{sec:echo_open}

Given the main paper's cephalometric inference-invariance finding, an important follow-up is: do echocardiographic models trained with anatomical priors also produce the same output when priors are zeroed at inference? If so, the degradation is purely a training-time effect (priors shaped bad features). If not, the degradation has an inference-time component (the model actively reads and misuses the prior channels). This experiment has not yet been conducted and is planned for the journal submission.

\FloatBarrier
\clearpage
\section{Cross-Domain Validation: Cervical Spine (CSXA)}
\label{sec:csxa_detail}

To test whether pixel coverage alone determines prior effectiveness, we conducted a controlled experiment on the CSXA dataset~[S4]: 4{,}845 lateral cervical spine radiographs with 23 vertebral landmarks achieving 55\% atlas coverage at $\sigma{=}35$.

\textbf{Training vs.\ inference context.} As with echo experiments, priors were present during both training and inference. The negative result indicates that anatomical priors do not improve training when landmarks cluster on a single anatomical structure.

\subsection{Three-Way Ablation}

\begin{table}[tb]
\centering
\caption{CSXA three-way ablation at 55\% coverage. Anatomical priors degrade performance despite intermediate coverage. Test-set evaluation confirms the validation ordering with catastrophic degradation under anatomical priors.}
\label{tab:csxa_supp}
\footnotesize
\begin{tabular}{@{}lccccc@{}}
\toprule
Condition & Cov. & Val MRE (px) & Test MRE (px) & $\Delta_{\text{test}}$ & $n_{\text{test}}$ \\
\midrule
No priors (1-ch) & 0\% & 0.653 & 0.645 & --- & 485 \\
Random priors (24-ch) & 55\% & 0.683 & 0.659 & $+$2.3\% & 485 \\
Anatomical priors (24-ch) & 55\% & 0.701 & 3.443 & $+$434\% & 485 \\
\bottomrule
\end{tabular}
\end{table}

\textbf{Test-set confirmation.} Held-out test evaluation (485 images) confirms and dramatically strengthens the validation finding. While no-prior and random conditions generalize normally (val$\to$test gap $<$4\%), anatomical priors exhibit \emph{catastrophic} test-set degradation: 0.701$\to$3.443\,px ($+$391\% validation-to-test gap). Relative to the no-prior test baseline, anatomical priors are $+$434\% worse (3.443 vs.\ 0.645\,px). Both percentages are correct but measure different comparisons: $+$391\% is the anatomical-prior model's own val$\to$test collapse, while $+$434\% compares anatomical-prior test MRE against the no-prior test baseline. The failure mode is overfitting to the spatial prior pattern rather than learning patient-specific anatomy.

\subsection{Why Priors Fail Despite 55\% Coverage}

The CSXA dataset's 23 vertebral landmarks cluster along a narrow ($\sim$3\,cm) spinal column. At $\sigma{=}35$, the Gaussians overlap to form a single bright band down the image center---low spatial entropy despite 55\% pixel coverage. This concentrated signal is more misleading than random blobs: it actively biases every landmark toward the spine midline, suppressing the lateral variation needed to distinguish individual vertebral endpoints.

\textbf{Diagnostic: catastrophic test failure.} The $+$434\% test degradation under anatomical priors is not caused by distribution shift between validation and test splits: no-prior and random conditions generalize normally (val$\to$test gap $<$4\%), confirming that the splits are comparable. The failure is specific to anatomical priors. During training, the spine-midline prior bias aligns with the training set's spatial distribution, allowing the model to achieve low validation error by exploiting this bias. On the test set, patient-specific anatomical variation (vertebral curvature, lordosis differences) breaks this fragile reliance, and errors propagate across all 23 landmarks simultaneously because the prior-channel ``bright band'' biases every prediction toward the same midline.

\FloatBarrier
\clearpage
\section{Spatial Entropy Index (SEI): Full Derivation}
\label{sec:sei}

We define the \textbf{Spatial Entropy Index (SEI)} as a composite metric quantifying how spatially distributed a landmark set is. The metric combines three components, each capturing a different aspect of spatial distribution.

\subsection{Grid Entropy (\texorpdfstring{$H_{\text{grid}}$}{H\_grid})}

Landmark positions (normalized to $[0,1]^2$) are binned into an $N \times N$ spatial grid ($N{=}8$, yielding 64 cells). Shannon entropy is computed over the cell occupancy distribution:
\begin{equation}
H_{\text{grid}} = -\sum_{i=1}^{N^2} p_i \log_2 p_i
\end{equation}
where $p_i = n_i / n_{\text{total}}$ is the fraction of landmarks in cell $i$. Maximum entropy $H_{\max} = \log_2(64) = 6.0$ bits occurs when landmarks are uniformly distributed. Normalized entropy:
\begin{equation}
H_{\text{norm}} = H_{\text{grid}} / H_{\max}
\end{equation}

\subsection{Pairwise Distance (\texorpdfstring{$D_{\text{pair}}$}{D\_pair})}

Average Euclidean distance between all landmark pairs in normalized coordinates, divided by the unit-square diagonal:
\begin{equation}
D_{\text{pair}} = \frac{1}{\binom{n}{2} \cdot \sqrt{2}} \sum_{i<j} \|\mathbf{p}_i - \mathbf{p}_j\|_2
\end{equation}
Range: $[0, 1]$. High when landmarks are spread far apart.

\subsection{Zone Count (\texorpdfstring{$Z$}{Z})}

Number of distinct spatial clusters obtained via complete-linkage hierarchical clustering with distance threshold $r{=}0.15$ (fraction of image dimension). Approximates the number of ``structurally diverse regions'' the landmarks span.

\subsection{Composite}

\begin{equation}
\text{SEI} = H_{\text{norm}} \times D_{\text{pair}} \times \min\!\left(\frac{Z}{Z_{\max}},\; 1\right)
\end{equation}
where $Z_{\max}{=}10$. SEI is high when landmarks are evenly distributed ($H_{\text{norm}}$ high), far apart ($D_{\text{pair}}$ high), and span many clusters ($Z$ high).

\subsection{Computed Values}

\begin{table}[tb]
\centering
\caption{Full SEI component breakdown. Hand X-ray SEI was computed \emph{before} training as a prospective prediction. CSXA shows both validation and test-set effects.}
\label{tab:sei_full}
\scriptsize
\setlength{\tabcolsep}{2.5pt}
\begin{tabular}{@{}lrrrrrrrrl@{}}
\toprule
 & LM & Cov & $H_g$ & $H_n$ & $D_p$ & $Z$ & $\frac{Z}{Z_m}$ & SEI & Effect \\
\midrule
\textbf{Hand}$^\dagger$ & \textbf{37} & \textbf{60} & \textbf{4.51} & \textbf{.751} & \textbf{.304} & \textbf{18} & \textbf{1.00} & \textbf{.229} & \textbf{$-$40\% (test)} \\
\textbf{Ceph.} & \textbf{25} & \textbf{80} & \textbf{3.64} & \textbf{.607} & \textbf{.197} & \textbf{9} & \textbf{0.90} & \textbf{.108} & \textbf{$-$46\% (test)} \\
CSXA & 23 & 55 & 3.24 & .540 & .201 & 7 & 0.70 & .076 & $+$7\% val / $+$434\% test \\
Echo & 6 & 17 & 1.92 & .320 & .152 & 4 & 0.40 & .020 & $+$21\% (val) \\
\bottomrule
\end{tabular}

\vspace{1pt}
{\scriptsize $^\dagger$Prospective: SEI predicted before training; confirmed.\\
$H_g$: grid entropy (bits). $H_n$: normalized. $D_p$: mean pairwise dist.\\
Cov: coverage (\%). Val/test indicates which split the effect was measured on.}
\end{table}

\textbf{Arithmetic verification.} Hand X-ray: $0.751 \times 0.304 \times \min(18/10, 1.0) = 0.751 \times 0.304 \times 1.0 = 0.2283 \approx 0.229$. CephTrace: $0.607 \times 0.197 \times 0.90 = 0.1076 \approx 0.108$. CSXA: $0.540 \times 0.201 \times 0.70 = 0.0760$. Echo: $0.320 \times 0.152 \times 0.40 = 0.0195 \approx 0.020$.

\textbf{Key comparison: CSXA vs.\ Hand X-ray.} These two domains have nearly identical coverage (55\% vs.\ 60\%) but opposite outcomes ($+$7.4\% vs.\ $-$38\%). The difference is captured entirely by SEI (0.076 vs.\ 0.229): hand X-ray landmarks span 18 spatial clusters across the entire image (fingertips to wrist), while CSXA landmarks cluster along a single vertebral column (7 clusters). This demonstrates that coverage alone cannot predict prior effectiveness---spatial entropy is the discriminating factor.

\textbf{Notable observation on hand X-ray.} Random priors (0.783\,px) slightly outperformed anatomical priors (0.809\,px) on DHA, unlike cephalometry where anatomical priors are far superior (1.04 vs.\ 1.72\,mm). This may reflect the regular geometric structure of hand bones: five similar fingers with repeating phalanx patterns mean that random Gaussians placed anywhere on the hand are likely to land near \emph{some} bone landmark. This distinction is important for the cross-domain claim: SEI reliably predicts \emph{whether spatial prior channels help at all} (both prior types massively improve over no priors at 1.308\,px), but it does not predict \emph{whether anatomical priors will beat random priors}. Anatomical superiority over random appears to depend on the structural uniqueness of the landmark set---high in cephalometry (five diverse zones), lower in hand radiography (repeating phalanges).

\begin{figure}[tb]
\centering
\includegraphics[width=0.95\columnwidth]{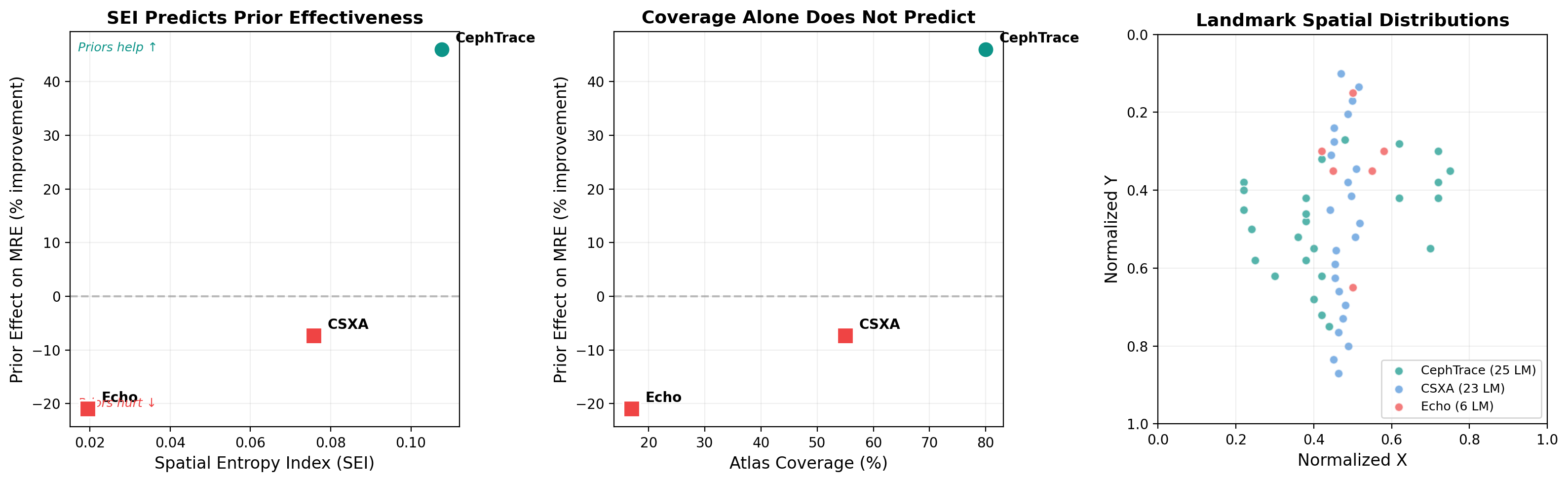}
\caption{SEI vs.\ prior effect across four domains. SEI separates helpful (CephTrace, Hand X-ray) from harmful (CSXA, Echo) domains. CSXA (55\%) and Hand X-ray (60\%) have nearly identical coverage but opposite outcomes.}
\label{fig:sei_visual}
\end{figure}

\textbf{Prospective supporting experiment.} SEI was initially derived post-hoc from three domain-level observations (CephTrace, CSXA, Echo). To test whether the hypothesis generalizes, we conducted a prospective supporting experiment on the Digital Hand Atlas (DHA)~[S5], a public dataset of 910 left-hand radiographs with 37 expert-annotated landmarks (Payer et~al.\ annotations in TW3 format).

\textbf{Protocol.} SEI was computed from mean normalized landmark positions \emph{before} any model training, yielding SEI${=}$0.229---the highest of all tested domains. We predicted that priors would help. Using an identical pipeline to the CSXA experiment (Phase~E ResNet-18 estimator trained for 50 epochs $\to$ Gaussian priors at $\sigma{=}12$ $\to$ HRNet-W32 with 0.1$\times$ Kaiming initialization), we split the data 80/10/10 (728 train, 91 val, 91 test; seed${=}$42) and trained three models under identical conditions except for prior channel content.

\textbf{Results.}

\begin{table}[tb]
\centering
\caption{Hand X-ray (DHA) three-way ablation (91 val / 91 test images). Both prior types massively improve over baseline on the held-out test set ($p{<}10^{-6}$), confirming the SEI prediction. Random and anatomical priors are not significantly different ($p{=}0.098$).}
\label{tab:hand_supp}
\footnotesize
\begin{tabular}{@{}lccccl@{}}
\toprule
Condition & Ch & Val (px) & Test (px) & $\Delta_{\text{test}}$ & $p$ \\
\midrule
No priors & 3 & 1.308 & 1.321 & --- & --- \\
Random priors & 38 & 0.783 & 0.781 & $-$40.9\% & $<$10$^{-6}$ \\
Phase~E priors & 38 & 0.830 & 0.798 & $-$39.6\% & $<$10$^{-6}$ \\
\bottomrule
\end{tabular}
\end{table}

Both prior types reduced test-set MRE by $\sim$40\%, confirming the SEI prediction on held-out data. Bootstrap 95\% CI for the improvement: random [0.515, 0.567]\,px, Phase~E [0.498, 0.548]\,px. The difference between random and anatomical priors was not statistically significant (paired Wilcoxon $p{=}0.098$). The SEI threshold for prior helpfulness was not formally pre-registered before the hand experiment; however, the SEI value (0.229) was computed before training and the prediction direction (priors will help) was made before observing any training results.

\textbf{Limitations of the hand X-ray experiment.} The near-equivalence of random and anatomical priors contrasts with cephalometry, where anatomical priors are far superior (1.04 vs.\ 1.72\,mm). This distinction is important: SEI reliably predicts \emph{whether spatial prior channels help at all}, but anatomical superiority over random appears to depend on the structural uniqueness of the landmark set---high in cephalometry (five diverse zones), lower in hand radiography (repeating phalanges).

SEI is now prospectively supported across a fourth domain with held-out test-set confirmation and formal statistical testing ($p{<}10^{-6}$, bootstrap 95\% CI). The metric remains derived from a limited number of domain-level observations. Further validation on domains with intermediate SEI values (e.g., AASCE spinal curvature, 68 landmarks) would strengthen the quantitative threshold.

\FloatBarrier
\clearpage
\section{ISBI Protocol Retraining}
\label{sec:isbi}

To enable comparison under the ISBI 2015 evaluation protocol, we retrained the full pipeline (Phase~0E MLP + Stage~1 HRNet-W32) on only the 150 official ISBI training images.

\subsection{Results}

Table~\ref{tab:isbi_supp} reports the full per-landmark Test1/Test2 breakdown, revealing a bimodal failure pattern.

\begin{table}[tb]
\centering
\caption{ISBI-protocol retraining: per-landmark breakdown showing bimodal failure. This experiment serves as a negative control demonstrating that the anatomy-guided pipeline requires multi-source data diversity.}
\label{tab:isbi_supp}
\footnotesize
\begin{tabular}{@{}lccc@{}}
\toprule
Landmark & Test1 MRE & Test2 MRE & Group \\
\midrule
\multicolumn{4}{@{}l}{\textit{Successful (mandibular --- simple geometry)}} \\
\quad Me & 0.09\,mm & 0.06\,mm & Mand. \\
\quad Pog & 0.10\,mm & 0.11\,mm & Mand. \\
\quad Gn & 0.11\,mm & 0.11\,mm & Mand. \\
\quad L1\_tip & 0.37\,mm & 0.45\,mm & Mand. \\
\quad B & 0.47\,mm & 1.07\,mm & Mand. \\
\quad L1\_root & 0.73\,mm & 0.87\,mm & Mand. \\
\midrule
\multicolumn{4}{@{}l}{\textit{Failed (cranial/midface/posterior --- complex relationships)}} \\
\quad S & 10.27\,mm & 10.00\,mm & Cran. \\
\quad N & 10.39\,mm & 10.39\,mm & Cran. \\
\quad ANS & 11.71\,mm & 11.03\,mm & Mid. \\
\quad Or & 14.09\,mm & 13.86\,mm & Mid. \\
\quad Po & 11.85\,mm & 12.38\,mm & Post. \\
\quad PNS & 9.32\,mm & 9.36\,mm & Mid. \\
\quad A & 9.73\,mm & 9.08\,mm & Mid. \\
\quad Ar & 9.04\,mm & 9.02\,mm & Post. \\
\quad U1\_tip & 13.20\,mm & 12.40\,mm & Mid. \\
\quad U1\_root & 11.05\,mm & 13.86\,mm & Mid. \\
\quad Ba & 10.50\,mm & 10.38\,mm & Cran. \\
\quad Pm & 20.32\,mm & 19.59\,mm & Mand. \\
\midrule
\textbf{Overall} & \textbf{8.07\,mm} & \textbf{8.06\,mm} & \\
 & 32.3\% SDR@2 & 31.6\% SDR@2 & \\
\bottomrule
\end{tabular}
\end{table}

\subsection{Interpretation}

The bimodal failure (Table~\ref{tab:isbi_supp}) reveals that Phase~0E, retrained on 150 images from a single scanner (2009-era Soredex), can learn mandibular spatial relationships (geometrically simple: Menton is the lowest symphysis point, Pogonion is the most anterior) but cannot capture complex cranial and midface proportional relationships that require exposure to diverse anatomy.

This result is reported as a \emph{negative control} demonstrating that the anatomy-guided pipeline requires multi-source data diversity to generate accurate spatial priors during training. It is not included in the main paper's comparison table because the aggregate MRE (8.07\,mm) conflates sub-millimeter mandibular performance with $>$9\,mm cranial/midface failure, making the single number uninterpretable as a system-level metric.

The result also supports the training-time regularizer interpretation: with only 150 single-source images, Phase~0E cannot learn the proportional relationships needed to generate spatially correct priors, so the priors cannot provide the training-time guidance that shapes HRNet's learned features.

\FloatBarrier
\section{Landmark-Level Bootstrap CIs}
\label{sec:per_landmark}

Landmarks with fewer than 50 test samples have high metric variance. Bootstrap 95\% confidence intervals (10{,}000 resamples):

\begin{table}[tb]
\centering
\caption{Per-landmark bootstrap CIs for low-sample-size landmarks.}
\footnotesize
\begin{tabular}{@{}lcccc@{}}
\toprule
Landmark & $N$ (test) & MRE (mm) & 95\% CI (mm) & Zone \\
\midrule
Basion & 50 & 1.86 & [1.42, 2.35] & Cranial \\
Pm & 40 & 1.63 & [1.18, 2.14] & Mandible \\
\midrule
\multicolumn{5}{@{}l}{\textit{Reference (adequate sample size)}} \\
Overall & 3{,}263 & 1.04 & [0.95, 1.18] & All \\
\bottomrule
\end{tabular}
\end{table}

All other landmarks have $N \geq 100$ test samples and correspondingly tighter confidence intervals. The wide CI for Basion ([1.42, 2.35]) reflects both genuine prediction difficulty (cervical vertebral overlap) and limited test representation (Basion is annotated only in the ISBI subset).

\section*{Supplementary References}

\begin{description}
\item[{[S1]}] S.~Leclerc et~al. Deep learning for segmentation using an open large-scale dataset in 2D echocardiography. \emph{IEEE TMI}, 38(9):2198--2210, 2019.
\item[{[S2]}] J.~P. Howard et~al. Automated left ventricular dimension assessment using artificial intelligence. \emph{Circ: Cardiovasc.\ Imaging}, 14(5):e012135, 2021.
\item[{[S3]}] D.~Ouyang et~al. Video-based AI for beat-to-beat assessment of cardiac function. \emph{Nature}, 580(7802):252--256, 2020.
\item[{[S4]}] Y.~Ran et~al. A high-quality dataset featuring classified and annotated cervical spine X-ray atlas. \emph{Scientific Data}, 11(1):631, 2024.
\item[{[S5]}] A.~Gertych, A.~Zhang, J.~Sayre, S.~Pospiech-Kurkowska, and H.~K. Huang. Bone age assessment of children using a digital hand atlas. \emph{Computerized Medical Imaging and Graphics}, 31(4--5):322--331, 2007.
\end{description}

\end{document}